\documentclass[a4paper]{article}
\usepackage[a4paper,top=3cm,left=3cm,right=2cm,bottom=2cm]{geometry}
\usepackage{amsmath}
\usepackage{amsthm}
\usepackage{graphicx}
\usepackage{natbib}
\usepackage{algorithm}
\usepackage{setspace}
\usepackage{fancyhdr}
\usepackage{amssymb}
\usepackage{setspace}
\usepackage[usenames,dvipsnames]{xcolor}

\usepackage{lscape}
\usepackage{graphics}
\usepackage{graphicx}
\usepackage{caption}
\usepackage{subcaption}
\usepackage{algpseudocode}
\usepackage{longtable}
\usepackage{multirow}

\newcommand{\up}[1]{\raisebox{1.3ex}[0pt]{#1}}

\begin{document}




\large
\title{Efficient local search limitation strategy for single machine total weighted tardiness scheduling with sequence-dependent setup times}

\author{{\bf Anand Subramanian, Katyanne Farias}\\
Departamento de Engenharia de Produ\c{c}\~ao\\ Universidade Federal da Para{\'i}ba, Brazil\\
anand@ct.ufpb.br, katyannefaraujo@gmail.com 
}
\date{}

\maketitle

\vspace{-0.5cm}
\begin{center}
Working Paper, UFPB -- November 2015 \\
\end{center}
\vspace{0.5cm}

\begin{abstract}

This paper concerns the single machine total weighted tardiness scheduling with sequence-dependent setup times, usually referred as $1|s_{ij}|\sum w_jT_j$. In this $\mathcal{NP}$-hard problem, each job has an associated processing time, due date and a weight. For each pair of jobs $i$ and $j$, there may be a setup time before starting to process $j$ in case this job is scheduled immediately after $i$. The objective is to determine a schedule that minimizes the total weighted tardiness, where the tardiness of a job is equal to its completion time minus its due date, in case the job is completely processed only after its due date, and is equal to zero otherwise. Due to its complexity, this problem is most commonly solved by heuristics. The aim of this work is to develop a simple yet effective limitation strategy that speeds up the local search procedure without a significant loss in the solution quality. Such strategy consists of a filtering mechanism that prevents unpromising moves to be evaluated. The proposed strategy has been embedded in a local search based metaheuristic from the literature and tested in classical benchmark instances. Computational experiments revealed that the limitation strategy enabled the metaheuristic to be extremely competitive when compared to other algorithms from the literature, since it allowed the use of a large number of neighborhood structures without a significant increase in the CPU time and, consequently, high quality solutions could be achieved in a matter of seconds. In addition, we analyzed the effectiveness of the proposed strategy in two other well-known metaheuristics. Further experiments were also carried out on benchmark instances of problem $1|s_{ij}|\sum T_j$.


  
\end{abstract}

%
%
%
\onehalfspace


\section{Introduction}
\label{sec:introduction}

This paper deals with the single machine total weighted tardiness scheduling with sequence-dependent setup times, a well-known problem in the scheduling literature, which can be defined as follows. Given a set of jobs $J = \{1,\dots,n\}$ to be scheduled on a single machine, for each job $j \in J$, let $p_j$ be the processing time and $d_j$ be the due date with a non-negative weight $w_j$. Also, consider a setup time $s_{ij}$ that is required before starting to process job $j \in J$ in case this job is scheduled immediately after job $i \in J$. The objective is to determine a schedule that minimizes the total weighted tardiness $\sum w_jT_j$, where the tardiness $T_j$ of a job $j \in J$ depends on its associated completion time $C_j$ and is given by $\max\{C_j - d_j, 0\}$. Based on the notation proposed in \cite{Graham1979} we will hereafter denote this problem as $1|s_{ij}|\sum w_jT_j$.

A special version of problem $1|s_{ij}|\sum w_jT_j$ arises when setup times are not considered. Such version is usually referred to as $1||\sum w_jT_j$, which is known to be $\mathcal{NP}$-hard in a strong sense \cite{Lawler1977, Lenstra1977}. Therefore, problem $1|s_{ij}|\sum w_jT_j$ is also $\mathcal{NP}$-hard since it includes problem $1||\sum w_jT_j$ as a particular case.

Given the complexity of problem $1|s_{ij}|\sum w_jT_j$, most methods proposed in the literature are based on (meta)heuristics. In particular, local search based metaheuristics had been quite effective in generating high quality solutions for this problem \citep{Kirlik2012, Xu2013, Subramanian2014}. Nevertheless, one of the main limitations of such methods is the computational cost of evaluating a move during the local search. Typically, the number of possible moves in classical neighborhoods such as insertion and swap is $O(n^2)$, whereas the complexity of evaluating each move from these neighborhoods is $O(n)$, when performed in a straightforward fashion. Hence, the overall complexity of examining these neighborhoods is $O(n^3)$.

Recently, \citet{Liaoetal2012} presented a sophisticated method that reduces the complexity of enumerating and evaluating all moves from the aforementioned neighborhoods from $O(n^3)$ to $O(n^2logn)$. However, in practice, this procedure is only useful  for very large instances. For example, they showed empirically that the advantage of using their approach for the neighborhood swap starts to be visibly significant only for instances with more than around 1000 jobs.

The aim of this work is to develop a simple yet effective strategy that speeds up the local search procedure without a significant loss in the solution quality. Such strategy consists of a filtering mechanism that prevents unpromising moves to be evaluated. The idea behind this approach relies on a measurement that must be computed at runtime. In our case, we use the setup variation, which can be computed in $O(1)$ time  to estimate if a move is promising or not to be evaluated. Moreover, the bottleneck of traditional implementations limits the number of neighborhoods to be explored in the local search. Our proposed strategy enables the use of a large number of neighborhoods, such as moving blocks of consecutive jobs, which is likely to lead to high quality solutions. In fact, \citet{Xu2013} showed that this type of neighborhood may indeed lead to better solutions for problem $1|s_{ij}|\sum w_jT_j$.  


The proposed local search strategy has been tested in classical benchmark instances using the ILS-RVND metaheuristic \citep{Subramanian2012b}. Computational experiments revealed that the limitation strategy enabled the metaheuristic to be extremely competitive when compared to other algorithms from the literature, since it allowed the use of a large number of neighborhood structures without a significant increase in the CPU time and, consequently, high quality solutions could be achieved in a matter of seconds. In addition, we analyzed the effectiveness of the proposed strategy in two other well-known metaheuristics. Further experiments were also carried out on benchmark instances of problem $1|s_{ij}|\sum T_j$.

The remainder of the paper is organized as follows. Section \ref {sec:literature} presents a review of the algorithms proposed in the literature for solving the problem $1|s_{ij}|\sum w_jT_j$. Section \ref{sec:strategy} describes the proposed local search methodology, including the neighborhood structures and the new limitation strategy. Section \ref{sec:experiments} presents the computational experiments. Section \ref{sec:conclusions} contains the concluding remarks of this work.

\section{Literature Review}
\label{sec:literature}

One of the first methods proposed for problem $1|s_{ij}|\sum w_jT_j$ was that of \citet{Raman1989}. It consists of a constructive heuristic based on a static dispatching rule. \citet{Lee1997} later developed a three-phase method, where the first performs a statistical analysis on the instance to define the parameters to be used, the second is a constructive heuristic that is based on dynamic dispatching rule called Apparent Tardiness Cost with Setups (ATCS), while the third performs a local search by means of insertion and swap moves.

The $1|s_{ij}|\sum w_jT_j$ literature remained practically unchanged for nearly two decades until \citet{Cicirello2005} developed five randomized based (meta)heuristic approaches, more precisely, Limited Discrepancy Search (LDS); Heuristic-Biased Stochastic Sampling (HBSS); Value-Biased Stochastic Sampling (VBSS); hill-climbing incorporated to VBSS (VBSS-HC) and finally a Simulated Annealing (SA) algorithm. The authors also proposed a set of 120 instances that has become the most used benchmark dataset for the problem. \citet{Cicirello2006} later implemented a Genetic Algorithm (GA) with a new operator called Non-Wrapping Order Crossover (NWOX) that maintains not only the relative order of the jobs in a sequence, but also the absolute one. 

Three metaheuristics were implemented by \citet{Lin2007}, more specifically, SA, GA and Tabu Search (TS), whereas Ant Colony Optimization (ACO) based algorithms were put forward by \citet{Liao2007}, \citet{Anghinolfi2008} and \citet{Mandahawi2011}. The latter authors actually developed a variant of ACO, called Max-Min Ant System (MMAS), that was capable of generating  better solutions when compared to the other two.

\citet{Valente2008} developed a Beam Search (BS) algorithm, while a Discrete Particle Swarm Optimization (DSPO) heuristic was proposed by \citet{Anghinolfi2009}. \citet{Tasgetiren2009} put forward a Discrete Differential Evolution (DDE) algorithm that was enhanced by the NEH constructive heuristic of \citet{Nawaz1983} combined with concepts of the metaheuristic Greedy Randomized Adaptive Search Procedure (GRASP) \citep{FeoResende1995} and some priority rules such as the Earliest Weighted Due Date (EWDD) and ATCS. The authors also implemented destruction and construction procedures to determine a mutant population.
\citet{Bozejko2010} proposed a parallel Scatter Search (ParSS) algorithm combined with a Path-Relinking scheme. \citet{Chao2012} implemented a so-called Discrete Electromagnetism-like Mechanism (DEM), which is a metaheuristic based on the electromagnetic theory of attraction and repulsion. 

\citet{Kirlik2012} developed a Generalized Variable Neighborhood Search (GVNS) algorithm combined with a Variable Neighborhood Descent (VND) procedure for the local search composed of the following neighborhoods: swap, 2-block insertion and job (1-block) insertion. \citet{Xu2013} suggested an Iterated Local Search (ILS) approach that uses a $l$-block insertion neighborhood structure with $l \in \{1, 2, \dots, 18\}$. The authors empirically showed that this neighborhood was capable of finding better solutions when compared to job insertion and 2-block insertion. The perturbation mechanism also applies the same type of move, but with $l \in \{18, 19, \dots, 30\}$.

\citet{Subramanian2014} also implemented an ILS algorithm, but combined with a Randomized VND procedure (RVND) that uses the neighborhoods insertion, 2-block insertion, 3-block insertion, swap and block reverse (a.k.a. twist). Diversification moves are performed using the double-bridge perturbation \citep{Martin1991}, which was originally developed for the Traveling Salesman Problem (TSP). An alternative version of the proposed ILS-RVND algorithm that accepts solutions with same cost during the local search was suggested. In this latter approach, a tabu list is used to avoid cycling. 

\citet{Deng2014} put forward an Enhanced Iterated Greedy (EIG) algorithm  that generates an initial solution using the ATCS heuristic, performs local search with swap and insertion moves and perturbs a solution by applying successive insertion moves. Some elimination rules were also suggested for the swap neighborhood. 

\citet{Xu2014} proposed different versions of a Hybrid Evolutionary Algorithm (HEA) by combining two population updating strategies with three crossover operators, including the linear order crossover operator (LOX). The initial population is generated at random and a local search is applied using the neighborhood $l$-block as in \cite{Xu2013}. The version that yielded the best results was denoted LOX$\oplus$B. \citet{Guo2015} suggested a Scatter Search (SS) based algorithm that combines many ideas from different methods such as ATCS, VNS and DDE with a new adaptive strategy for updating the reference set.  

To the best of our knowledge, the ILS-RVND heuristic of \citet{Subramanian2014}, the ILS of \citet{Xu2013} and the HEA of \citet{Xu2014}  are the best algorithms, at least in terms of solution quality, proposed for problem $1|s_{ij}|\sum w_jT_j$.

\citet{Tanaka2013} proposed an exact algorithm, called Successive Sublimation Dynamic Programming (SSDP), that was capable of solving instances of the benchmark dataset of \citet{Cicirello2005}. At first, a column generation or a conjugate subgradient procedure is applied over the lagrangean relaxation of the original problem, solved by dynamic programming. Next, some constraints are added to the relaxation until there is no difference between the lower and upper bounds, which increases the number of states in the dynamic programming. Unnecessary states are removed in order to decrease the computational time and the memory used. A branching scheme is integrated to the method to solve the harder instances. Despite capable of solving all instances to optimality, the computational time was, on average, rather large, more specifically 32424.55 seconds, varying from 0.54 seconds to 30 days.

Table \ref{Summary} shows a summary of the methods proposed for problem $1|s_{ij}|\sum w_jT_j$. For the sake of comparison we also included, when applicable, the neighborhoods used in each method. It can be observed that most algorithms used the neighborhoods insertion and swap. Moreover, it is interesting to notice that almost all population based algorithms rely on local search at a given step of the method.

\begin{table}[!ht]
  \centering
  \onehalfspacing
   \footnotesize
 \setlength{\tabcolsep}{2.0mm}
  \caption{Summary of the methods proposed for problem $1|s_{ij}|\sum w_jT_j$}
\begin{tabular}{cccc}
    \hline 
    Work & Year & Method & Neighborhoods used \\ 
    \hline
	 \citet{Raman1989} & 1989 & Constructive heuristic & -- \\ [3pt]
	 \citet{Lee1997} & 1997 & Constructive heuristic & -- \\ [3pt]
	 & & LDS, HBSS, VBSS & \\ 
	 \up{\citet{Cicirello2005}} & \up{2005} & VBSS-HC, SA & \up{Not reported}\\ [3pt]
	 \citet{Cicirello2006} & 2006 & GA & -- \\ [3pt]
	 \citet{LiaoJuan2007} & 2007 & ACO & Insertion, swap \\ [3pt]
	 \citet{Lin2007} & 2007 & GA, SA, TS & Insertion, swap \\ [3pt]
	  &  & & Insertion$^1$, 2-block swap,  \\ 
	 \up{\citet{Valente2008}} & \up{2008} & \up{BS} & 3-block swap \\ [3pt]
	 \citet{Anghinolfi2008} & 2008 & ACO & Insertion, swap \\ [3pt]
	 \citet{Anghinolfi2009} & 2009 & DPSO & Insertion, swap \\ [3pt]
	 \citet{Tasgetiren2009} & 2009 & DDE & Insertion \\ [3pt]
	 \citet{Bozejko2010} & 2010 & ParSS & Swap \\ [3pt]
	 \citet{Mandahawi2011} & 2011 & ACO & Insertion \\ [3pt]
	 \citet{Chao2012} & 2012 & DEM & Insertion$^2$ \\ [3pt]
	  &  &  & Insertion, \\ 
	 \up{\citet{Kirlik2012}} & \up{2012} & \up{GVNS} &  2-block insertion, swap \\ [3pt]
	 \citet{Tanaka2013} & 2013 & SSDP (exact) & -- \\ [3pt]
	 \citet{Xu2013} & 2013 & ILS & $l$-block insertion$^3$ \\ [3pt]
	 \citet{Deng2014} & 2014 & EIG & insertion, swap \\ [3pt]
			     & 		& & Insertion, 2-block insertion,  \\
	 \up{\citet{Subramanian2014}} & \up{2014} & \up{ILS-RVND} & 3-Insertion, swap, block reverse \\ [3pt]
	 \citet{Xu2014} & 2014 & HEA & $l$-block insertion$^3$ \\ [3pt]
	 \citet{Guo2015} & 2015 & SS & Insertion, 2-block insertion, swap, 3-opt \\ 
	 \hline
	 \multicolumn{4}{l}{\scriptsize $^1$: Choose the job with the largest weighted tardiness and insert that job after $n/3$ jobs.}\\ 	 
	 \multicolumn{4}{l}{\scriptsize $^2$: Random insertion, insertion of the best job in the best position, insertion of a random job in the best position.} \\
	 \multicolumn{4}{l}{\scriptsize $^3$: With $l \in \{1, 2,\dots, 18\}$.}
	 
\end{tabular}
\label{Summary}
\end{table}

\section{Local Search Methodology}
\label{sec:strategy}

In this section we describe in detail the local search methodology adopted in this work. At first,  we introduce the proposed local search limitation strategy used to speed up the local search process. Next, we present the neighborhood structures considered in our implementation and how we compute the move evaluation using a block representation. Finally, we show how we have embedded the developed approach into the ILS-RVND metaheuristic \citep{Subramanian2012b}.

\subsection{Limitation Strategy}
\label{sec:limitation}

Enumerating and evaluating all moves of a given neighborhood of a $1|s_{ij}|\sum w_jT_j$ solution are usually very time consuming, often being the bottleneck of local search based algorithms proposed for this problem. 
\citet{Liaoetal2012} developed a rather complicated procedure that runs in $O(n^2logn)$ time for preprocessing the auxiliary data stuctures necessary for performing the move evaluation of the neighborhoods swap, insertion and block reverse in constant time. Nevertheless, they themselves showed, for example, that for the neighborhood swap, their approach only start to be notably superior for instances containing more than $\approx 1000$ jobs. 
Therefore, it was thought advisable to use the simple and straightforward move evaluation procedures in Section \ref{sec:neighborhoods}, even though their resulting overall complexity is $O(n^3)$. However, instead of enumerating all possible moves from each neighborhood, we decided to evaluate only a subset of them, selected by means of a novel local search limitation strategy.

Our proposed local search limitation strategy is based on a very simple filtering mechanism that efficiently chooses the neighboring solutions to be evaluated during the search. The criterion used to decide weather a move should be evaluated or not is based on the setup variation of that move, which can be computed in $O(1)$ time. In other words, for every potential move, one computes the setup variation and the move is only considered for evaluation if the variation does not exceed a given threshold value. The motivation for adopting such criterion was based on the empirical observation that the larger the increase on the total setup of a sequence due to a move, the smaller the probability of improvement on the total weighted tardiness. 

It is worthy of note that, very recently, \citet{Guo2015} had independently developed a similar but not identical approach to disregard unpromising moves based on the total setup time instead of the setup variation. More specifically, moves are not considered for evaluation if the total setup time of a solution to be evaluated is greater than a threshold value that is computed by multiplying the average setup of the instance by a random number selected from the interval between 0.2 and 0.3. However, they did not report any result of the impact of this strategy on the performance of the local search.

Let $\mathcal{N}$ be the set of neighborhoods used in a local search algorithm. A threshold value $max \Delta s_v$ for the setup variation is assigned for each neighborhood $v \in \mathcal{N}$, meaning that a move from a neighborhood $v \in \mathcal{N}$ will only be evaluated if its associated setup variation is smaller than or equal to $max \Delta s_v$. 
Since an adequate value for each  $max \Delta s_v$ may be highly sensitive to the instance data as well as the neighborhood, we did not impose any predetermined value for them. Instead, they are estimated during a learning phase, when a preliminary local search is performed without any filter.

The learning phase occurs during the first iterations of the local search. Initially, $max \Delta s_v, \forall v \in \mathcal{N}$, is set to a sufficiently large number $M$ so as to allow the evaluation of any move, i.e., no filter is applied at this stage. Let $\Delta s_{v}, \forall v \in \mathcal{N}$, be a list composed of the values of the setup variations associated to each improving move of a particular neighborhood. This list is updated with the addition of a new value every time an improving move occurs. 

At the end of the learning phase, the list $\Delta s_{v}, \forall v \in \mathcal{N}$, is sorted in ascending order. Define $\theta$ as an input parameter, where $0 \leq \theta \leq 1$.  The element from this ordered list associated to the position $\lfloor{\theta \cdot |\Delta s_v|\rfloor}$ is then chosen as threshold value for the parameter $max \Delta s_{v}, \forall v \in \mathcal{N}$. Note that the larger the $\theta$, the larger the $max \Delta s_{v}$ and thus more moves are likely to be evaluated, leading to a more conservative limitation policy.

Consider an example where $\Delta s_{swap} = [-6, -4, -4, -2, 0, 1, 4, 7, 12, 20]$. If $\theta = 0.95$, then $\lfloor{\theta \cdot |\Delta s_v|\rfloor} = \lfloor{0.95 \cdot 10\rfloor} = 9$. The value of the parameter $max \Delta s_{swap}$ will thus be the one associated with the position 9 of $\Delta s_{swap}$, that is, $max \Delta s_{swap} = 12$. 

Note that one need not necessarily perform an exhaustive local search, i.e. enumerate all possibles moves from a neighborhood during the learning phase. What is really relevant is the size of the list $\Delta s_v$. On one hand, if the size of the sample ($|\Delta s_v|$) is too small and not really representative, then an inaccurate value for $max \Delta s_v$ will be estimated. On the other hand, storing a large sample may imply in spending a considerable amount of iterations in the learning phase, which can dramatically affect the performance of the algorithm in terms of CPU time.


The presented limitation strategy can be easily embedded into any local search based metaheuristic. For example, in case of multi-start metaheuristics such as GRASP, one can consider the learning phase (i.e., perform the search without any filters) only for a given number of preliminary iterations before triggering the limitation scheme. In case of other metaheuristics that systematically alternate between intensification and diversification such as TS, VNS and ILS, one can choose different stopping criteria for the learning phase such as the number of calls to the local search procedure or even a time limit. Population based metaheuristics that rely on local search for obtaining good solutions can also employ a similar scheme.

\subsection{Neighborhood structures}
\label{sec:neighborhoods}

\begin{figure}[!ht]
 \begin{center}
  \includegraphics[scale=0.15]{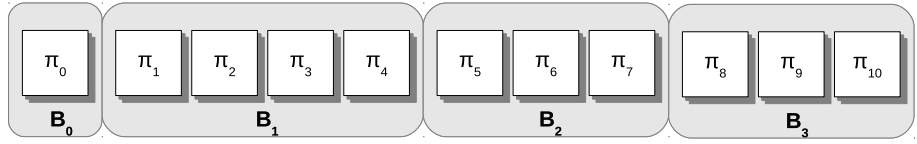}\\
  \caption{Example of a schedule divided into 4 blocks}
\label{fig:blocks}
 \end{center}  
\end{figure}

In order to better describe the neighborhood structures employed during the local search we use the following block representation. Let $\pi = \{\pi_0,\pi_1, \pi_2,\dots,\pi_n\}$ be an arbitrary sequence composed of $n$ jobs, where $\pi_{0} = 0$ is a dummy job that is used for considering the setup $s_{0\pi_1}$ required for processing the first job of the sequence.  A given block $B$ can be defined as a subsequence of consecutive jobs. Figure \ref{fig:blocks} shows an example of a sequence of 10 jobs (plus the dummy job) divided into 4 blocks, namely $B_0 = \pi_0$, $B_1 = \{\pi_1,\pi_2,\pi_3,\pi_4\}$, $B_2 = \{\pi_5,\pi_6,\pi_7\}$ and $B_3 = \{\pi_8,\pi_9,\pi_{10}\}$. 

Let $a$ and $b$ be the position of the first and last jobs of a block $B_t$ in the sequence, respectively, and let $b'$ be the position of the last job of the predecessor block $B_{t-1}$ in the sequence with associated completion time $C\_b'$. The cost of the block $B_t$, i.e., its total weighted tardiness, can be computed in $O(|B_t|$) steps as shown in Alg. \ref{alg:CompCostBlock}. Note that if $|B_t| = 1$, the procedure will not execute the loop from $a+1$ to $b$ described in lines \ref{loop-start}-\ref{loop-end} because in this case $a = b$ which implies in $b < a + 1$. 

\begin{algorithm}[!ht]
    \caption{CompCostBlock}	 
\label{alg:CompCostBlock}
\footnotesize
\begin{algorithmic}[1]
\State \textbf{Procedure} CompCostBlock($b', a, b, \pi, C\_b'$) 
\State $cost \leftarrow 0$
\State $Ctemp \leftarrow C\_b' + s_{\pi_{b'}\pi_{a}} + p_{\pi_{a}}$ \scriptsize{{\color{gray}\Comment{Global variable that stores a temporary completion time}}}
\If {$Ctemp > d_{\pi_{a}}$} \scriptsize{{\color{gray}\Comment{$d_{\pi_{a}}$ is the due date of job $\pi_{a}$}}}
    \State $cost \leftarrow w_{\pi_{a}} \times (Ctemp - d_{\pi_{a}})$ \scriptsize{{\color{gray}\Comment{$w_{\pi_{a}}$ is the weight of job $\pi_{a}$}}}
\EndIf 
\For {$a' = a+1 \dots b$} \label{loop-start}
    \State $Ctemp \leftarrow Ctemp + s_{\pi_{a'-1}\pi_{a'}} + p_{\pi_{a'}}$ 
    \If {$Ctemp > d_{\pi_{a'}}$}
	\State $cost \leftarrow cost + w_{\pi_{a'}} \times (Ctemp - d_{\pi_{a'}})$
    \EndIf 
\EndFor \label{loop-end}
\State \Return $cost$
\State \textbf{end} CompCostBlock.
\end{algorithmic}
\end{algorithm}

It is easy to verify that the total cost of a sequence can be obtained by the sum of the costs of the blocks defined for that sequence. In the example given in Fig. \ref{fig:blocks} the total cost of the sequence is equal to the sum of the cost of blocks $B_0$, $B_1$, $B_2$ and $B_3$. 

When performing move evaluations it is quite useful to define an auxiliary data structure that stores the cumulated weighted tardiness up to a certain position of the sequence. Therefore, we have decided to define an array $g$ for that purpose. For example, the element $g_5$ of the array stores the cumulated weighted tardiness up to the 5th position of the sequence.

We now proceed to a detailed description of the neighborhood structures used in our approach.

\subsubsection{Swap}
\label{sec:swap}

\begin{figure}[!ht]
 \begin{center}
  \includegraphics[scale=0.15]{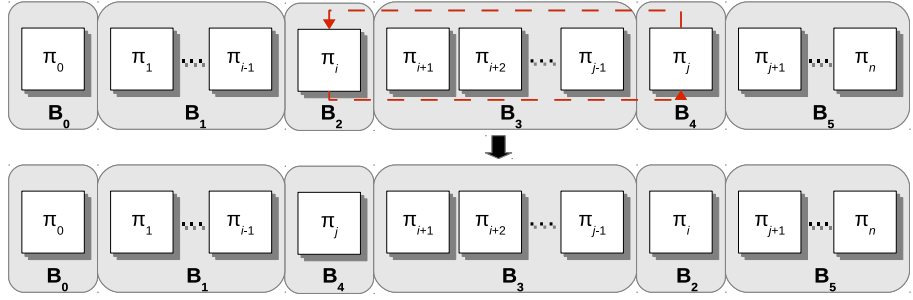}\\
  \caption{Exchanging the position of two jobs}
\label{fig:swap}
 \end{center}  
\end{figure}

The swap neighborhood structure simply consists of exchanging the position of two jobs in the sequence, as depicted in Figure \ref{fig:swap}. It is possible to observe that the modified solution can be divided into 6 blocks. Alg. \ref{alg:CompCostSwap} shows how to compute the cost of a solution in $O(n-i)$ steps. Note that one can always assume that $i < j$ for every swap move. 

\begin{algorithm}[!ht]
    \caption{CompCostSwap}	 
\label{alg:CompCostSwap}
\footnotesize
\begin{algorithmic}[1]
\State \textbf{Procedure} CompCostSwap($\pi, i, j, g, C$)
\State $f \leftarrow g_{i-1}$  {\color{gray}\Comment{Variable that stores the cost to be evaluated}}
\State $Ctemp \leftarrow C_{\pi_{i-1}}$ 
\State $f \leftarrow f + \texttt{CompCostBlock}(i-1, j, j, \pi, Ctemp)$ {\color{gray}\Comment{Cost of block 4}}
\State $f \leftarrow f + \texttt{CompCostBlock}(j, i+1, j-1, \pi, Ctemp)$ {\color{gray}\Comment{Cost of block 3}}
\State $f \leftarrow f + \texttt{CompCostBlock}(j-1, i, i, \pi, Ctemp)$ {\color{gray}\Comment{Cost of block 2}}
\State $f \leftarrow f + \texttt{CompCostBlock}(i, j+1, n, \pi, Ctemp)$ {\color{gray}\Comment{Cost of block 5}}
\State \Return $f$
\State \textbf{end} CompCostSwap. 
\end{algorithmic}
\end{algorithm}

Alg. \ref{alg:Swap} presents the pseudocode of the neighborhood swap considering the limitation strategy described in Section \ref{sec:limitation}. At first, the best sequence $\pi^*$ is considered to be the same as the original one, i.e., $\pi$ (line \ref{alg:Swap:update1}). Next, for every pair of positions $i$ and $j$ (with $j > i$) of $\pi$ (lines \ref{alg:Swap:for}-\ref{alg:Swap:endfor}), the setup variation is computed in constant time (line \ref{alg:Swap:setup}). If this variation is smaller than or equal to $max \Delta s_{swap}$  (lines \ref{alg:Swap:if}-\ref{alg:Swap:endif}), then the cost of sequence  $\pi'$, which is a neighbor of $\pi$ generated by exchanging the position of customers $\pi_i$ and $\pi_j$, is computed using Alg. \ref{alg:CompCostSwap} (line \ref{alg:Swap:cost}). In case of improvement (lines \ref{alg:Swap:if}-\ref{alg:Swap:endif2}), the best current solution is updated (line \ref{alg:Swap:update2}) and if the learning phase is activated, that is, if $max\Delta s_{swap}$ = $M$, 
list $\Delta s_{swap}$ is updated by adding the value associated to the setup variation computed in line \ref{alg:Swap:setup} (lines \ref{alg:Swap:if3}-\ref{alg:Swap:endif3}). Finally, the procedure returns the best sequence $\pi^*$ found during the search (line \ref{alg:Swap:return}).

\begin{algorithm}[ht]
    \caption{Swap}	 
\label{alg:Swap}
\footnotesize
\begin{algorithmic}[1]
\State {\bf Procedure} Swap($\pi, g, C, \Delta s_{swap}, max\Delta s_{swap}$)
\State $\pi^* \leftarrow \pi$; $f^* \leftarrow f(\pi)$; \label{alg:Swap:update1}		
\For {$i=1 \dots n-1$} \label{alg:Swap:for}
    \For {$j=i+1 \dots n$}
	\State $setupVariation = - s_{\pi_{i-1}\pi_{i}} - s_{\pi_{i}\pi_{i+1}} - s_{\pi_{j-1}\pi_{j}} - s_{\pi_{j}\pi_{j+1}}$ \label{alg:Swap:setup}
	\Statex \hspace{3.8cm} $+ s_{\pi_{i-1}\pi_{j}} + s_{\pi_{j}\pi_{i+1}} + s_{\pi_{j-1}\pi_{i}} + s_{\pi_{i}\pi_{j+1}}$
        \If {$setupVariation \leq max\Delta s_{swap}$} \label{alg:Swap:if}
	    \State $f(\pi')$ = \texttt{CompCostSwap}($\pi$, $i$, $j$, $l$, $g$, $C$) {\color{gray}\Comment{$\pi'$ is a neighbor of $\pi$}} \label{alg:Swap:cost}
		\If {$f(\pi') < f^*$} \label{alg:Swap:if2}
		\State $\pi^* \leftarrow \pi'$; $f^* \leftarrow f(\pi')$ \label{alg:Swap:update2}		
		\If {$max\Delta s_{swap}$ = $M$}  {\color{gray}\Comment{Learning phase activated}} \label{alg:Swap:if3}
		    \State $\Delta s_{swap} \leftarrow \Delta s_{swap} \cup setupVariation$ \label{alg:Swap:update3}
		\EndIf  \label{alg:Swap:endif3}
	    \EndIf \label{alg:Swap:endif2}
	\EndIf \label{alg:Swap:endif}
    \EndFor
\EndFor \label{alg:Swap:endfor}
\State \textbf{return} $\pi^*$; \label{alg:Swap:return}
\State \textbf{end} Swap.
\end{algorithmic}
\end{algorithm}

\subsubsection{$l$-block insertion}
\label{sec:l-block}

\begin{figure}[!ht]
 \begin{center}
  \includegraphics[scale=0.15]{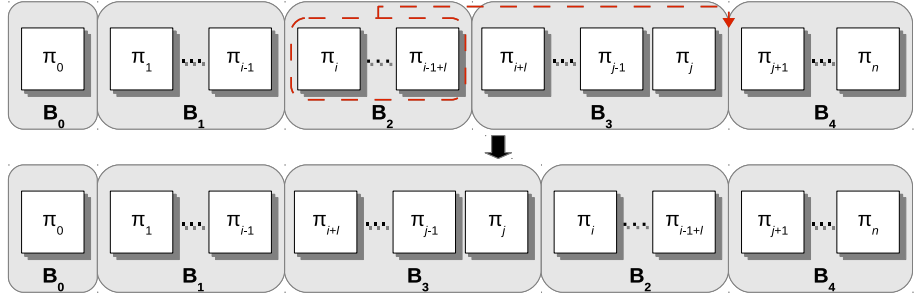}\\
  \caption{Forward insertion of a block of length $l$}
\label{fig:l-block-Fwd}
 \end{center}  
\end{figure}

\begin{figure}[!ht]
 \begin{center}
  \includegraphics[scale=0.15]{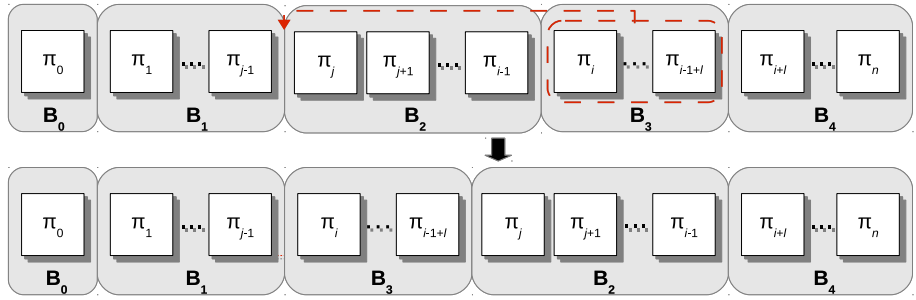}\\
  \caption{Backward insertion of a block of length $l$}
  \label{fig:l-block-Back}
 \end{center}  
\end{figure}

The $l$-block insertion neighborhood consists of moving a block of length $l$ forward ($i < j)$ or backward ($i > j$), as shown in Figs. \ref{fig:l-block-Fwd} and \ref{fig:l-block-Back}, respectively. In this case the resulting solution in both cases can be represented by 5 blocks. Algs. \ref{alg:CompCostFl-block} and \ref{alg:CompCostBl-block} describe how to evaluate a cost of moving a block forward and backward in $O(n-i)$ and $O(n-j)$ steps, respectively. 

\begin{algorithm}[!ht]
    \caption{CompCost$l$-blockF}	 
\label{alg:CompCostFl-block}
\footnotesize
\begin{algorithmic}[1]
\State \textbf{Procedure} CompCost$l$-blockF($\pi, i, j, l, g, C$)
\State $f \leftarrow g_{i-1}$  {\color{gray}\Comment{Variable that stores the cost to be evaluated}}
\State $Ctemp \leftarrow C_{\pi_{i-1}}$ 
\State $f \leftarrow f + \texttt{CompCostBlock}(i-1, i+l, j, \pi, Ctemp)$ {\color{gray}\Comment{Cost of block 3}}
\State $f \leftarrow f + \texttt{CompCostBlock}(j, i, i-1+l, \pi, Ctemp)$ {\color{gray}\Comment{Cost of block 2}}
\State $f \leftarrow f + \texttt{CompCostBlock}(i-1+l, j+1, n, \pi, Ctemp)$ {\color{gray}\Comment{Cost of block 4}}
\State \Return $f$
\State \textbf{end} CompCostF$l$-block. 
\end{algorithmic}
\end{algorithm}

\begin{algorithm}[!htb]
    \caption{CompCost$l$-blockB}	 
\label{alg:CompCostBl-block}
\footnotesize
\begin{algorithmic}[1]
\State \textbf{Procedure} CompCost$l$-blockB($\pi, i, j, l, g, C$)
\State $f \leftarrow g_{j-1}$  {\color{gray}\Comment{Variable that stores the cost to be evaluated}}
\State $Ctemp \leftarrow C_{\pi_{j-1}}$  
\State $f \leftarrow f + \texttt{CompCostBlock}(j-1, i, i-1+l, \pi, Ctemp)$ {\color{gray}\Comment{Cost of block 3}}
\State $f \leftarrow f + \texttt{CompCostBlock}(i-1+l, j, i-1, \pi, Ctemp)$ {\color{gray}\Comment{Cost of block 2}}
\State $f \leftarrow f + \texttt{CompCostBlock}(i-1, i+l, n, \pi, Ctemp)$ {\color{gray}\Comment{Cost of block 4}}
\State \Return $f$
\State \textbf{end} CompCostB$l$-block. 
\end{algorithmic}
\end{algorithm}

It is worth mentioning that one can check at any time if the partial cost of the solution under evaluation is already worse than the best current cost. If so, the evaluation can be interrupted because it is already known that the solution under evaluation cannot be better than the current one. In some cases this may help speeding up the move evaluation process.

Let $L$ be a set composed of different values for the parameter $l$. In practice one can define a single neighborhood considering all values of $l \in L$ at once (see \cite{Xu2013}), or define multiple neighborhoods where each of them is associated with a given value of $l \in L$. In our implementation we decided for the latter option. 


Alg. \ref{alg:Insertion_l-block} shows the pseudocode of the neighborhood $l$-block taking into account the limitation strategy described earlier. This algorithm, which is divided into two parts, follows the same rationale of Alg. \ref{alg:Swap} in both of them. The main difference is related to the range of the positions $i$ and $j$, which depends on the value of $l$. In part one (lines \ref{alg:Insertion_l-block:part1beg}-\ref{alg:Insertion_l-block:part1end}) the search is performed forward, whereas in part 2 it is performed backward (lines \ref{alg:Insertion_l-block:part2beg}-\ref{alg:Insertion_l-block:part2end}).

\begin{algorithm}[!ht]
    \caption{$l$-blockInsertion}	 
\label{alg:Insertion_l-block}
\footnotesize
\begin{algorithmic}[1]
\State {\bf Procedure} $l$-blockInsertion($\pi, l, g, C, \Delta s_{l\text{-}block}, max\Delta s_{l\text{-}block}$)
\State $\pi^* \leftarrow \pi$; $f^* \leftarrow f(\pi)$; 
\Statex {\color{gray}\Comment{Moving the $l$-block forward} \hspace{10.8cm}}  
\For {$i=1 \dots n-l$} \label{alg:Insertion_l-block:part1beg}
    \For {$j=i+1 \dots n$}
	\State $setupVariation = - s_{\pi_{i-1}\pi_{i}} - s_{\pi_{i-1+l}\pi_{i+l}} - s_{\pi_{j}\pi_{j+1}}$   
	\Statex \hspace{3.8cm} $ + s_{\pi_{i-1}\pi_{i+l}} + s_{\pi_{j}\pi_{i}} + s_{\pi_{i-1+l}\pi_{j+1}}$ 
        \If {$setupVariation \leq max\Delta s_{l\text{-}block}$}
	    \State $f(\pi')$ = \texttt{CompCostl-blockF}($\pi$, $i$, $j$, $l$, $g$, $C$) {\color{gray}\Comment{$\pi'$ is a neighbor of $\pi$}}
	    \If {$f(\pi') < f^*$}
		\State $\pi^* \leftarrow \pi'$; $f^* \leftarrow f(\pi')$		
		\If {$max\Delta s_{l\text{-}block}$ = $M$}  {\color{gray}\Comment{Learning phase activated}}
		    \State $\Delta s_{l\text{-}block} \leftarrow \Delta s_{l\text{-}block} \cup setupVariation$; 
		\EndIf
	    \EndIf
	\EndIf
    \EndFor
\EndFor \label{alg:Insertion_l-block:part1end}
\Statex {\color{gray}\Comment{Moving the $l$-block backward \hspace{10.5cm}}}
\For {$i=2 \dots n-l+1$} \label{alg:Insertion_l-block:part2beg}
  \For {$j=1 \dots i-1$}
	\State $setupVariation = - s_{\pi_{j-1}\pi_{j}} - s_{\pi_{i-1}\pi_{i}} - s_{\pi_{i-1+l}\pi_{i+1}}$ 
	\Statex \hspace{3.8cm} $+ s_{\pi_{j-1}\pi_{i}} + s_{\pi_{j-1+l}\pi_{j}} + s_{\pi_{i-1}\pi_{i+1}}$
        \If {$setupVariation \leq max\Delta s_{l\text{-}block}$}
	    \State $f(\pi')$ = \texttt{CompCostl-blockB}($\pi$, $i$, $j$, $l$, $g$, $C$) {\color{gray}\Comment{$\pi'$ is a neighbor of $\pi$}}
	    \If {$f(\pi') < f^*$}
		\State $\pi^* \leftarrow \pi'$; $f^* \leftarrow f(\pi')$
		\If {$max\Delta s_{l\text{-}block}$ = $M$} {\color{gray}\Comment{Learning phase activated}}
		    \State $\Delta s_{l\text{-}block} \leftarrow \Delta s_{l\text{-}block} \cup setupVariation$
		\EndIf
	    \EndIf
	\EndIf
    \EndFor
\EndFor \label{alg:Insertion_l-block:part2end}
\State \textbf{return} $\pi^*$;
\State \textbf{end} $l$-blockInsertion.
\end{algorithmic}
\end{algorithm}



\subsection{Embedding the proposed approach in the ILS-RVND metaheuristic}
\label{sec:ILS-RVND}

In this section we explain how we embedded the proposed approach in the ILS-RVND metaheuristic \citep{Subramanian2012b}. The reason for choosing this metaheuristic to test our local search limitation strategy is that it was found capable of generating highly competitive results not only for problem $1|s_{ij}|\sum w_jT_j$ \citep{Subramanian2014}, as mentioned in Section \ref{sec:literature}, but also for other combinatorial optimization problems \citep{Subramanian2010, Subramanian2012b, Silvaetal2012, Pennaetal2013, SubramanianBattarra2013, Martinellietal2013, Vidaletal2015}. Moreover, the referred method relies on very few parameters and it can be considered relatively simple, which makes it quite practical and also easy to implement.

Alg. \ref{alg:FastILSvsILS} highlights the differences between the original version of ILS-RVND and a new one, called ILS-RVND$_{Fast}$, that includes the additional steps (line \ref{alg:FILS:setupParam} and lines \ref{alg:FILS:if}-\ref{alg:FILS:endif}) described in Section \ref{sec:limitation} for speeding up the the local search. It can be observed that the learning phase is limited to the first iteration of the algorithm, where the list $\Delta s_v, \forall v \in \mathcal{N}$, is populated during the local search (line \ref{alg:FILS:LS}). The value of $max \Delta s_v, \forall v \in \mathcal{N}$ is then estimated after the end of the first iteration. Parameters $I_R$ and $I_{ILS}$ correspond to the number of restarts of the metaheuristic and the number of consecutive ILS iterations without improvements, respectively. An exception occurs during the learning phase, which is more costly, where the number of ILS iterations is $I_{ILS}/2$. In both algorithms the initial solution (line \ref{alg:FILS:const}) is 
generated using a simple randomized insertion heuristic, whereas the perturbation (line \ref{alg:FILS:pertub}) is performed by a mechanism called double-bridge, which consists of exchanging two blocks of a sequence at random. The reader is referred to \cite{Subramanian2014} for implementation details about these procedures.

The local search method of both algorithms is performed by a RVND procedure, which consists of selecting an unexplored neighborhood at random whenever another one fails to find an improved solution. In case of improvement, all neighborhoods are reconsidered to be explored.  Note that in both algorithms the set $L$ associated to the $l$-block neighborhood is provided as an input parameter to be tuned (see Section \ref{sec:tuning}), as opposed to the ILS-RVND presented in \cite{Subramanian2014} where this set was predefined as $L = \{1, 2, 3\}$, that is, the $l$-block neighborhoods were limited to 1-, 2- and 3-block insertion. In addition, the neighborhood block reverse was also used in a restricted fashion by the authors, but it was not considered here because it did not seem to be crucial for obtaining high quality solutions. 

One last remark is that the algorithm stops when a sequence $\pi$ with cost $f(\pi) = 0$ is found. When this happens it is clear that an optimal solution was obtained and thus there is no point in continuing the search. This was not originally considered in the ILS-RVND presented in \cite{Subramanian2014}, but it has been considered here in both versions.

\begin{landscape}
\begin{algorithm}[!ht]
    \caption{ILS-RVND vs ILS-RVND$_{Fast}$}
\begin{minipage}{0.60\textwidth}
\label{alg:FastILSvsILS}
\footnotesize
\begin{algorithmic}[1]
\State \textbf{Procedure} ILS-RVND($I_{R}, I_{ILS}, L$)
\State $f^{*}$ $\leftarrow$ $\infty$
\State \text{}
\For {$iter=1 \dots I_{R}$} \label{alg:ILS:loop-start1}
    \State $\pi \leftarrow \texttt{GenerateInitialSolution}()$ \label{alg:ILS:const}
    \State $\hat{\pi} \leftarrow \pi$
    \State $iterILS\leftarrow 0$
    \While {$iterILS \leq I_{ILS}$}   \label{alg:ILS:loop-start2}	
        \State $\pi \leftarrow \texttt{RVND}(\pi, L)$ \label{alg:ILS:LS}
        \If {$f(\pi) < f(\hat{\pi}$)}
            \State $\hat{\pi} \leftarrow \pi$
            \State $iterILS \leftarrow 0$
        \EndIf
        \State $\pi \leftarrow \texttt{Perturb}(\hat{\pi})$ \label{alg:ILS:pertub}
        \State $iterILS \leftarrow iterILS + 1$
    \EndWhile \label{alg:ILS:loop-end2}
    \State
    \State 
    \State
    \State
    \State 
    \State
    \If {$f(\hat{\pi}) < f^{*}$}
        \State $\pi^{*} \leftarrow \hat{\pi}$; $f^{*}$ $\leftarrow$ $f(\hat{\pi})$
    \EndIf
\EndFor \label{alg:ILS:loop-end1}
\State \textbf{return} $\pi^{*}$ \label{alg:ILS:return}
\State \textbf{end} ILS-RVND.
\end{algorithmic}
\end{minipage}
\hspace*{0.2cm}
\begin{minipage}{0.76\textwidth}
\label{alg:F-ILS-RVND}
\footnotesize
\begin{algorithmic}
\State \textbf{Procedure} ILS-RVND$_{Fast}$($I_{R}, I_{ILS}, \theta, L$)
\State $f^{*}$ $\leftarrow$ $\infty$
\State $max\Delta s_{v} \leftarrow M; \Delta s_v \leftarrow \texttt{NULL}, \forall v \in \mathcal{N}$ \label{alg:FILS:setupParam} 
\For {$iter=1 \dots I_{R}$} \label{alg:FILS:loop-start1}
    \State $\pi \leftarrow \texttt{GenerateInitialSolution}()$ \label{alg:FILS:const}
    \State $\hat{\pi} \leftarrow \pi$
    \State $iterILS \leftarrow 0$
    \While {$iterILS \leq I_{ILS}$} \label{alg:FILS:loop-start2} {\color{gray}\Comment{or $iterILS \leq I_{ILS}/2$ when $iter = 0$}} 
        \State $\pi \leftarrow \texttt{RVND}(\pi, L, \Delta s, max\Delta s)$ \label{alg:FILS:LS}
        \If {$f(\pi) < f(\hat{\pi}$)}
            \State $\hat{\pi} \leftarrow \pi$
            \State $iterILS \leftarrow 0$
        \EndIf
        \State $\pi \leftarrow \texttt{Perturb}(\hat{\pi})$ \label{alg:FILS:pertub}
        \State $iterILS \leftarrow iterILS + 1$
    \EndWhile \label{alg:FILS:loop-end2}
    \If {$iter = 1$} \label{alg:FILS:if}
        \For {$v = 1 \dots |\mathcal{N}|$}
            \State $\Delta s_v \leftarrow$ \texttt{sort}($\Delta s_v$) \label{alg:FILS:listS}
            \State {$max\Delta s_{v} \leftarrow$ value associated to the position $\lfloor\theta\cdot|\Delta s_v|\rfloor$ of list $\Delta s_v$} \label{alg:FILS:maxS}
        \EndFor
    \EndIf \label{alg:FILS:endif}
    \If {$f(\hat{\pi}) < f^{*}$}
        \State $\pi^{*} \leftarrow \hat{\pi}$; $f^{*}$ $\leftarrow$ $f(\hat{\pi})$
    \EndIf
\EndFor \label{alg:FILS:loop-end1}
\State \textbf{return} $\pi^{*}$ \label{alg:FILS:return}
\State \textbf{end} ILS-RVND$_{Fast}$.
\end{algorithmic}
\end{minipage}
\end{algorithm}
\end{landscape}

\section{Computational Experiments}
\label{sec:experiments}

The ILS-RVND and ILS-RVND$_{Fast}$ algorithms were coded in C++ and the experiments were performed in an Intel Core i7 with 3.40 GHz and 16 GB of RAM running under Linux Mint 13. Only a single thread was used in our testing.

The 120 instances of \citet{Cicirello2005} were used to evaluate the performance of the proposed algorithms. Each of them has 60 jobs and is characterized by three parameters: $\tau$, which is related to the tightness of the due date; $R$, which specifies the range of the due dates; $\eta$, which refers to the size of the average setup time with respect to the size of the average processing time. The authors created 12 groups of 10 instances by combining the following parameters: $\tau = \{0.3, 0.6, 0.9\}$, $R = \{0.25, 0.75\}$ and $\eta = \{0.25, 0.75\}$, as shown in Table \ref{Instances}.

\begin{table}[!ht]
  \centering
  \onehalfspacing
   \footnotesize
  \caption{Group of instances generated by \citet{Cicirello2005} for problem $1|s_{ij}|\sum w_jT_j$}
\begin{tabular}{ccc}
    \hline 
    Group & Instances & Configuration \\ 
    \hline
    1	& 1-10	& $\tau = 0.3, R = 0.25, \eta = 0.25$ \\  
    2	& 11-20	& $\tau = 0.3, R = 0.25, \eta = 0.75$ \\  
    3	& 21-30	& $\tau = 0.3, R = 0.75, \eta = 0.25$ \\  
    4	& 31-40	& $\tau = 0.3, R = 0.75, \eta = 0.75$ \\      
    5	& 41-50	& $\tau = 0.6, R = 0.25, \eta = 0.25$ \\  
    6	& 51-60	& $\tau = 0.6, R = 0.25, \eta = 0.75$ \\  
    7	& 61-70	& $\tau = 0.6, R = 0.75, \eta = 0.25$ \\     
    8	& 71-80	& $\tau = 0.6, R = 0.75, \eta = 0.75$ \\      
    9	& 81-90	& $\tau = 0.9, R = 0.25, \eta = 0.25$ \\  
    10	& 91-100&$\tau = 0.9, R = 0.25, \eta = 0.75$ \\  
    11	& 101-110&$\tau = 0.9, R = 0.75, \eta = 0.25$ \\  
    12	& 111-120&$\tau = 0.9, R = 0.75, \eta = 0.75$ \\  
    \hline	 
\end{tabular}
\label{Instances}
\end{table}

\subsection{Parameter Tuning}
\label{sec:tuning}

To have a better idea of the impact of the modifications introduced in the original ILS-RVND, we decided to use the same configuration for the number of ILS iterations as in \cite{Subramanian2014}, that is, $I_{ILS} = 4 \times n$. However, since each iteration of the algorithm became much faster after implementing the limitation strategy (see Section \ref{sec:impact}) we decided to set $I_{R} = 20$, instead of $I_{R} = 10$, as in \cite{Subramanian2014}, because it seemed to provide a better compromise between solution quality and computational time.

A set of 17 challenging instances was selected for tuning the parameters $L$ and $\theta$, namely instances 1, 2, 3, 4, 5, 7, 8. 9, 10, 11, 13, 14, 15, 16, 18, 20 and 24. The criterion used for choosing these instances was based on the difficulty faced by the ILS-RVND implemented in \cite{Subramanian2014} in finding their corresponding optimal solutions.

Let \emph{Best Gap} be the gap between the best solution found in 10 runs and the optimal solution; \emph{Worst Gap} be the gap between the worst solution found in 10 runs and the optimal solution; and \emph{Avg. Gap} be the average gap between the average solution of 10 runs and the optimal solution. In the tables presented hereafter, Arithm. Mean of Best Gaps (\%) corresponds to the arithmetic mean of the \emph{Best Gaps}, Geom. Mean of Avg. Gaps (\%) denotes the geometric mean of the \emph{Avg. Gaps}, Geom. Mean of Worst Gaps (\%) indicates the geometric mean of the \emph{Worst Gaps} and Arithm. Mean of Avg. Time (s) is the arithmetic mean of the average times in seconds. Given a set of $q$ positive numbers, the geometric mean can be defined as the $q$th root of the product of all numbers of the set. The reason for using the geometric mean is because it normalizes the different ranges, that is, the possible significant differences among the Gaps. Hence, the geometric mean was adopted in some of the cases 
as an attempt to perform a fair comparison between the solutions found for each configuration tested. However, we used the arithmetic mean for the \emph{Best Gaps} because in many cases the best solution found coincided with the optimal solution. In this case, the gap is equal to zero and thus the value is disregarded from the computation of the geometric mean, which may result in a misleading result when many values are not considered. 

We have tested 34 different combination of neighborhoods, more precisely, we considered $L = \{1,\dots,4\}$ to $L = \{1,\dots, 20\}$ incrementing one $l$-block neighborhood at a time and we tried each possibility with and without swap. Therefore, the goal of this experiment was not only to calibrate the parameter $L$, but also to investigate the benefits of including the neighborhood swap. For this testing, we have arbitrarily set $\theta = 0.90$ and we ran the algorithm 10 times for each instance. Table \ref{L} shows the average results obtained for the 17 instances mentioned above with the different combinations of neighborhoods. We decided to adopt the configuration $L  = \{1,\dots,13\}$ + swap because it seemed to offer a good balance between solution quality and computational time.

\begin{table}[!htbp]
\caption{Results for the parameter tuning of $L$ with and without the neighborhood swap}
\centering
\scriptsize
\onehalfspacing
\begin{tabular}{lcccc}
\hline
\multicolumn{ 1}{c}{} & Arithm. Mean of & Geom. Mean of & Geom. Mean of & Arithm. Mean of   \\ 
\multicolumn{ 1}{c}{\up{Neighborhoods}} & Best Gaps (\%) & Avg. Gaps (\%)  &  Worst Gaps (\%) & Avg. Times (s)   \\ \hline
$L  = \{1,\dots,4\}$  & 2.95 & 2.94 & 5.02 & 6.44 \\ 
$L  = \{1,\dots,4\}$ + swap & 3.23 & 2.76 & 4.93 & 6.12 \\ 
$L  = \{1,\dots,5\}$  & 2.29 & 2.07 & 4.12 & 6.66 \\ 
$L  = \{1,\dots,5\}$ + swap & 2.43 & 1.94 & 3.22 & 6.96 \\ 
$L  = \{1,\dots,6\}$  & 1.95 & 1.44 & 3.01 & 7.66 \\ 
$L  = \{1,\dots,6\}$ + swap & 1.44 & 1.38 & 2.95 & 7.51 \\ 
$L  = \{1,\dots,7\}$  & 1.07 & 1.13 & 2.60 & 15.58 \\ 
$L  = \{1,\dots,7\}$ + swap & 1.72 & 1.07 & 2.05 & 7.95 \\ 
$L  = \{1,\dots,8\}$  & 0.82 & 1.16 & 2.61 & 12.75 \\ 
$L  = \{1,\dots,8\}$ + swap & 2.05 & 1.18 & 1.98 & 8.22 \\ 
$L  = \{1,\dots,9\}$  & 1.42 & 1.09 & 2.21 & 10.11 \\ 
$L  = \{1,\dots,9\}$ + swap & 1.24 & 0.90 & 1.68 & 8.68 \\ 
$L  = \{1,\dots,10\}$  & 1.92 & 0.91 & 1.79 & 10.96 \\ 
$L  = \{1,\dots,10\}$ + swap & 1.11 & 0.79 & 1.74 & 9.23 \\ 
$L  = \{1,\dots,11\}$  & 1.05 & 1.00 & 1.84 & 12.15 \\ 
$L  = \{1,\dots,11\}$ + swap & 0.75 & 0.88 & 1.70 & 9.72 \\ 
$L  = \{1,\dots,12\}$  & 0.30 & 0.79 & 1.64 & 9.47 \\ 
$L  = \{1,\dots,12\}$ + swap & 0.51 & 0.79 & 1.70 & 9.73 \\ 
$L  = \{1,\dots,13\}$  & 1.06 & 0.80 & 1.94 & 10.02 \\ 
$\textbf{\textit{L}  = \{1,\dots,13\}}$ \textbf{+ swap} & \textbf{0.12} & \textbf{0.74} & \textbf{1.39} & \textbf{9.85} \\ 
$L  = \{1,\dots,14\}$  & 1.03 & 0.68 & 1.58 & 10.33 \\ 
$L  = \{1,\dots,14\}$ + swap & 0.39 & 0.84 & 1.69 & 10.24 \\ 
$L  = \{1,\dots,15\}$  & 1.15 & 0.78 & 1.66 & 10.40 \\ 
$L  = \{1,\dots,15\}$ + swap & 1.35 & 0.71 & 1.49 & 10.42 \\ 
$L  = \{1,\dots,16\}$  & 1.01 & 0.89 & 1.81 & 10.70 \\ 
$L  = \{1,\dots,16\}$ + swap & 0.31 & 0.70 & 1.49 & 11.00 \\ 
$L  = \{1,\dots,17\}$  & 0.81 & 0.78 & 1.56 & 11.29 \\ 
$L  = \{1,\dots,17\}$ + swap & 0.75 & 0.80 & 1.52 & 11.39 \\
$L  = \{1,\dots,18\}$  & 0.83 & 0.77 & 1.60 & 11.72 \\ 
$L  = \{1,\dots,18\}$ + swap & 0.31 & 0.71 & 1.57 & 11.63 \\ 
$L  = \{1,\dots,19\}$  & 0.13 & 0.74 & 1.53 & 11.74 \\ 
$L  = \{1,\dots,19\}$ + swap & 0.90 & 0.90 & 1.83 & 11.81 \\ 
$L  = \{1,\dots,20\}$  & 1.32 & 0.93 & 1.72 & 12.16 \\ 
$L  = \{1,\dots,20\}$ + swap & 1.07 & 0.74 & 1.41 & 11.94 \\ \hline
\end{tabular}
\label{L}
\end{table}

Five different values for the parameter $\theta$ was tested, in particular, 0,80, 0.85, 0.90, 0.95 and 1.00, and the results can be found in Table \ref{theta}. As expected, it can be observed that the average computational time is directly proportional to  $\theta$ since more moves are considered for evaluation as the value of $\theta$ increases. Note that $\theta = 1.00$ implies that $max\Delta s_{v}$ is equal to the largest setup variation associated to an improving move of neighborhood $v \in \mathcal{N}$ that occurred during the learning phase. We decided to adopt $\theta = 0.90$ because solutions of similar quality were obtained when compared to $\theta = 1.00$, but approximately four times faster. 

\begin{table}[!htbp]
\caption{Results for the parameter tuning of $\theta$}
\centering
\scriptsize
\onehalfspacing
\begin{tabular}{ccccc}
\hline
\multicolumn{ 1}{c}{} & Arithm. Mean of & Geom. Mean of & Geom. Mean of & Arithm. Mean of   \\ 
\multicolumn{ 1}{c}{\up{$\theta$}} & Best Gaps (\%) & Avg. Gaps (\%)  &  Worst Gaps (\%) & Avg. Times (s)   \\ \hline
0.80 & 0.15 & 0.97 & 2.05 & 7.44 \\ 
0.85 & 1.25 & 0.78 & 1.56 & 8.11 \\ 
\textbf{0.90} & \textbf{1.11} & \textbf{0.75} & \textbf{1.59} & \textbf{10.12} \\ 
0.95 & 1.51 & 0.66 & 1.34 & 16.81 \\ 
1.00 & 0.91 & 0.74 & 1.70 & 41.74 \\ \hline
\end{tabular}
\label{theta}
\end{table}

\subsection{Comparison with the literature}
\label{sec:comparison}

In this section we compare the best, average and worst results found by ILS-RVND$_{Fast}$ over 10 runs with the best methods available in the literature. We specify below the algorithms considered for comparison as well as the type of result reported by the associated work.

\textbf{ACO$_{\text{AP}}$}: Ant Colony Optimization of \citet{Anghinolfi2008}. Best of 10 runs.

\textbf{DPSO}: Discrete Particle Swarm Optimization of \citet{Anghinolfi2009}. Best of 10 runs.

\textbf{DDE}: Discrete Differential Evolutionary heuristic of \citet{Tasgetiren2009}. Best of 10 runs.

\textbf{GVNS}: General Variable Neighborhood Search of \citet{Kirlik2012}. Best of 20 runs.

\textbf{ILS$_{\text{XLC}}$}: Iterated Local Search of \citet{Xu2013}. Best and average of 100 runs.

\textbf{ILS-RVND$_{\text{SBP}}$}: Iterated Local Search + Randomized Variable Neighborhood Descent of \citet{Subramanian2014}. Best and average of 10 runs.

\textbf{LOX$\oplus$B}: Hybrid Evolutionary Algorithm of \citet{Xu2014}. Best and average of 20 runs.

\textbf{Opt}: Optimal solution found by the exact algorithm of \citet{Tanaka2013}.\\

Table \ref{Comparison} presents the results found by ILS-RVND$_{Fast}$ and also by the algorithms mentioned above, except ACO$_{\text{AP}}$ for space restriction reasons. The average times reported in this table are in seconds.  It can be observed that the proposed algorithm was capable of finding the optimal solution for all instances, except for instance 24. Moreover, it is possible to verify that the mean of the average times of ILS-RVND$_{Fast}$ was approximately 13 seconds.  

\begin{center}
{\scriptsize
\onehalfspacing
\setlength{\tabcolsep}{0.51mm}
\setlength{\LTcapwidth}{10in}
\begin{longtable}{ccccccccccccccccccccrr}
\caption{Results for the instances of \citet{Cicirello2005}} \\
\hline
\multicolumn{ 1}{c}{} & \multicolumn{ 1}{c}{} & $\!\!\!$DPSO$\!\!\!$ &  & $\!\!\!$DDE$\!\!\!$ &  & $\!\!\!$GVNS$\!\!\!$ &  & \multicolumn{ 2}{c}{$\!\!\!$ILS$_{\text{XLC}}\!\!\!$} &  & \multicolumn{ 2}{c}{$\!\!\!$ILS-RVND$_{\text{SBP}}\!\!\!$} &  & \multicolumn{2}{c}{$\!\!\!$LOX$\oplus$B$\!\!\!$} &  & \multicolumn{ 4}{c}{$\!\!\!$ILS-RVND$_{Fast}\!\!\!$} & \\ \cline{3-3}  \cline{5-5} \cline{7-7}  \cline{9-10} \cline{12-13} \cline{15-16} \cline{18-21}
\multicolumn{ 1}{c}{Inst} & \multicolumn{ 1}{c}{$\!\!\!$Opt$\!\!\!$} &  &  &  &  &  &  &  &  &  &  &  &  &  &  &  & & &  & Avg.  \\ 
\multicolumn{ 1}{c}{} & \multicolumn{ 1}{c}{} &  \up{Best} &  & \up{Best} &  & \up{Best} &  & \up{Best} & \up{Avg.} &  & \up{Best} & \up{Avg.} &  &  \up{Best} & \up{Avg.} &  & \up{Best} & \up{Avg.} & \up{Worst} & Time  \\ 
\hline
\endfirsthead
\caption[]{(\textit{Continued})}\\
\hline
\multicolumn{ 1}{c}{} & \multicolumn{ 1}{c}{} & $\!\!\!$DPSO$\!\!\!$ &  & $\!\!\!$DDE$\!\!\!$ &  & $\!\!\!$GVNS$\!\!\!$ &  & \multicolumn{ 2}{c}{$\!\!\!$ILS$_{\text{XLC}}\!\!\!$} &  & \multicolumn{ 2}{c}{$\!\!\!$ILS-RVND$_{\text{SBP}}\!\!\!$} &  & \multicolumn{2}{c}{$\!\!\!$LOX$\oplus$B$\!\!\!$} &  & \multicolumn{ 4}{c}{$\!\!\!$ILS-RVND$_{Fast}\!\!\!$} & \\ \cline{3-3}  \cline{5-5} \cline{7-7}  \cline{9-10} \cline{12-13} \cline{15-16} \cline{18-21}
\multicolumn{ 1}{c}{Inst} & \multicolumn{ 1}{c}{$\!\!\!$Opt$\!\!\!$} &  &  &  &  &  &  &  &  &  &  &  &  &  &  &  & & &  & Avg.  \\ 
\multicolumn{ 1}{c}{} & \multicolumn{ 1}{c}{} &  \up{Best} &  & \up{Best} &  & \up{Best} &  & \up{Best} & \up{Avg.} &  & \up{Best} & \up{Avg.} &  &  \up{Best} & \up{Avg.} &  & \up{Best} & \up{Avg.} & \up{Worst} & Time  \\ 
\hline
\endhead
\hline \multicolumn{22}{r}{\textit{Continued on next page}} \\
\endfoot
\hline
\endlastfoot
1 & 453 & 531 &  & 474 &  & 471 &  & 453 & 480.3 &  & 459 & 470.4 &  & 453 & 462.2  &  & 453 & 457.5 & 459 & 8.5 \\ 
2 & 4794 & 5088 &  & 4902 &  & 4878 &  & 4794 & 4887.0 &  & 4866 & 4910.4 &  & 4794 & 4841.5  &  & 4794 & 4813.8 & 4842 & 11.1 \\ 
3 & 1390 & 1609 &  & 1465 &  & 1430 &  & 1390 & 1457.5 &  & 1414 & 1433.9 &  & 1390 & 1401.8  &  & 1390 & 1393 & 1395 & 10.8 \\ 
4 & 5866 & 6146 &  & 5946 &  & 6006 &  & 5866 & 5978.4 &  & 5906 & 5982 &  & 5866 & 5871.2  &  & 5866 & 5866 & 5866 & 7.0 \\ 
5 & 4054 & 4339 &  & 4084 &  & 4114 &  & 4074 & 4215.9 &  & 4084 & 4129 &  & 4054 & 4096.9  &  & 4054 & 4072 & 4084 & 10.9 \\ 
6 & 6592 & 6832 &  & 6652 &  & 6667 &  & 6592 & 6750.3 &  & 6607 & 6665.5 &  & 6592 & 6617.2  &  & 6592 & 6593.5 & 6607 & 9.1 \\ 
7 & 3267 & 3514 &  & 3350 &  & 3330 &  & 3267 & 3404.2 &  & 3350 & 3394.2 &  & 3267 & 3319.0  &  & 3267 & 3281.2 & 3296 & 14.2 \\ 
8 & 100 & 132 &  & 114 &  & 108 &  & 100 & 106.2 &  & 105 & 109.6 &  & 100 & 102.1  &  & 100 & 101 & 102 & 7.7 \\ 
9 & 5660 & 6153 &  & 5803 &  & 5751 &  & 5660 & 5840.8 &  & 5673 & 5760.1 &  & 5660 & 5699.5  &  & 5660 & 5665.2 & 5673 & 11.3 \\ 
10 & 1740 & 1895 &  & 1799 &  & 1789 &  & 1740 & 1793.0 &  & 1768 & 1783.5 &  & 1740 & 1756.8  &  & 1740 & 1746.3 & 1761 & 9.3 \\ 
11 & 2785 & 3649 &  & 3294 &  & 2998 &  & 2830 & 3125.0 &  & 2934 & 3062.8 &  & 2798 & 2871.0  &  & 2785 & 2858.1 & 2894 & 8.19 \\ 
12 & 0 & 0 &  & 0 &  & 0 &  & 0 & 0.0 &  & 0 & 0 &  & 0 & 0.0  &  & 0 & 0 & 0 & $< 0.1$ \\ 
13 & 3904 & 4430 &  & 4194 &  & 4068 &  & 3942 & 4141.9 &  & 4014 & 4103.1 &  & 3904 & 3986.5  &  & 3904 & 3949.0 & 3978 & 7.86 \\ 
14 & 2075 & 2749 &  & 2268 &  & 2260 &  & 2081 & 2315.1 &  & 2219 & 2279.4 &  & 2075 & 2179.3  &  & 2075 & 2128.2 & 2174 & 9.45 \\ 
15 & 724 & 1250 &  & 964 &  & 935 &  & 775 & 891.3 &  & 896 & 953.6 &  & 724 & 794.8  &  & 724 & 771 & 809 & 7.1 \\ 
16 & 3285 & 4127 &  & 3876 &  & 3381 &  & 3285 & 3413.5 &  & 3325 & 3415.3 &  & 3285 & 3306.6  &  & 3285 & 3296.2 & 3301 & 9.3 \\ 
17 & 0 & 75 &  & 61 &  & 0 &  & 0 & 13.6 &  & 0 & 31.1 &  & 0 & 2.5  &  & 0 & 0 & 0 & 1.2 \\ 
18 & 767 & 971 &  & 857 &  & 845 &  & 767 & 813.5 &  & 787 & 812.5 &  & 767 & 789.0  &  & 767 & 776.4 & 789 & 8.6 \\ 
19 & 0 & 0 &  & 0 &  & 0 &  & 0 & 0.0 &  & 0 & 0 &  & 0 & 0.0  &  & 0 & 0 & 0 & $< 0.1$ \\ 
20 & 1757 & 2675 &  & 2111 &  & 2053 &  & 1757 & 1938.9 &  & 1789 & 1920.2 &  & 1757 & 1790.9  &  & 1757 & 1757.8 & 1761 & 7.5 \\ 
21 & 0 & 0 &  & 0 &  & 0 &  & 0 & 0.0 &  & 0 & 0 &  & 0 & 0.0  &  & 0 & 0 & 0 & $< 0.1$ \\ 
22 & 0 & 0 &  & 0 &  & 0 &  & 0 & 0.0 &  & 0 & 0 &  & 0 & 0.0  &  & 0 & 0 & 0 & $< 0.1$ \\ 
23 & 0 & 0 &  & 0 &  & 0 &  & 0 & 0.0 &  & 0 & 0 &  & 0 & 0.0  &  & 0 & 0 & 0 & $< 0.1$ \\ 
24 & 761 & 1043 &  & 1033 &  & 920 &  & 773 & 1037.2 &  & 1004 & 1028 &  & 761 & 1027.8  &  & 773 & 967.9 & 1030 & 24.5 \\ 
25 & 0 & 0 &  & 0 &  & 0 &  & 0 & 0.0 &  & 0 & 0 &  & 0 & 0.0  &  & 0 & 0 & 0 & $< 0.1$ \\ 
26 & 0 & 0 &  & 0 &  & 0 &  & 0 & 0.0 &  & 0 & 0 &  & 0 & 0.0  &  & 0 & 0 & 0 & $< 0.1$ \\ 
27 & 0 & 0 &  & 0 &  & 0 &  & 0 & 0.0 &  & 0 & 0 &  & 0 & 0.0  &  & 0 & 0 & 0 & 0.2 \\ 
28 & 0 & 0 &  & 0 &  & 0 &  & 0 & 0.0 &  & 0 & 0 &  & 0 & 0.0  &  & 0 & 0 & 0 & $< 0.1$ \\ 
29 & 0 & 0 &  & 0 &  & 0 &  & 0 & 0.0 &  & 0 & 0 &  & 0 & 0.0  &  & 0 & 0 & 0 & $< 0.1$ \\ 
30 & 0 & 0 &  & 0 &  & 0 &  & 0 & 28.2 &  & 0 & 0 &  & 0 & 24.9  &  & 0 & 0 & 0 & 0.2 \\ 
31 & 0 & 0 &  & 0 &  & 0 &  & 0 & 0.0 &  & 0 & 0 &  & 0 & 0.0  &  & 0 & 0 & 0 & $< 0.1$ \\ 
32 & 0 & 0 &  & 0 &  & 0 &  & 0 & 0.0 &  & 0 & 0 &  & 0 & 0.0  &  & 0 & 0 & 0 & $< 0.1$ \\ 
33 & 0 & 0 &  & 0 &  & 0 &  & 0 & 0.0 &  & 0 & 0 &  & 0 & 0.0  &  & 0 & 0 & 0 & $< 0.1$ \\ 
34 & 0 & 0 &  & 0 &  & 0 &  & 0 & 0.0 &  & 0 & 0 &  & 0 & 0.0  &  & 0 & 0 & 0 & $< 0.1$ \\ 
35 & 0 & 0 &  & 0 &  & 0 &  & 0 & 0.0 &  & 0 & 0 &  & 0 & 0.0  &  & 0 & 0 & 0 & $< 0.1$ \\ 
36 & 0 & 0 &  & 0 &  & 0 &  & 0 & 0.0 &  & 0 & 0 &  & 0 & 0.0  &  & 0 & 0 & 0 & $< 0.1$ \\ 
37 & 0 & 186 &  & 107 &  & 46 &  & 0 & 293.3 &  & 0 & 119 &  & 0 & 77.4  &  & 0 & 18.0 & 56 & 16.8 \\ 
38 & 0 & 0 &  & 0 &  & 0 &  & 0 & 0.0 &  & 0 & 0 &  & 0 & 0.0  &  & 0 & 0 & 0 & $< 0.1$ \\ 
39 & 0 & 0 &  & 0 &  & 0 &  & 0 & 0.0 &  & 0 & 0 &  & 0 & 0.0  &  & 0 & 0 & 0 & $< 0.1$ \\ 
40 & 0 & 0 &  & 0 &  & 0 &  & 0 & 0.0 &  & 0 & 0 &  & 0 & 0.0  &  & 0 & 0 & 0 & $< 0.1$ \\ 
41 & 69102 & 69102 &  & 69242 &  & 69242 &  & 69102 & 69318.2 &  & 69102 & 69200 &  & 69102 & 69222.7  &  & 69102 & 69102 & 69102 & 21.2 \\ 
42 & 57487 & 57487 &  & 57511 &  & 57511 &  & 57487 & 57644.6 &  & 57487 & 57489.4 &  & 57487 & 57558.0  &  & 57487 & 57487 & 57487 & 17.5 \\ 
43 & 145310 & 145883 &  & 145310 &  & 145310 &  & 145310 & 145930.5 &  & 145310 & 145771.7 &  & 145310 & 145659.6  &  & 145310 & 145370 & 145794 & 15.3 \\ 
44 & 35166 & 35331 &  & 35289 &  & 35289 &  & 35166 & 35328.3 &  & 35166 & 35251.4 &  & 35166 & 35253.8  &  & 35166 & 35166 & 35166 & 13.0 \\ 
45 & 58935 & 59175 &  & 58935 &  & 59025 &  & 58935 & 59018.6 &  & 58935 & 58944 &  & 58935 & 58974.6  &  & 58935 & 58935 & 58935 & 19.7 \\ 
46 & 34764 & 34805 &  & 34764 &  & 34764 &  & 34764 & 35034.8 &  & 34764 & 34817.3 &  & 34764 & 34896.8  &  & 34764 & 34764 & 34764 & 18.6 \\ 
47 & 72853 & 73378 &  & 73005 &  & 72853 &  & 72853 & 73160.4 &  & 72853 & 73028 &  & 72853 & 73070.6  &  & 72853 & 72853 & 72853 & 16.6 \\ 
48 & 64612 & 64612 &  & 64612 &  & 64612 &  & 64612 & 64719.6 &  & 64612 & 64638.9 &  & 64612 & 64699.6  &  & 64612 & 64612 & 64612 & 30.2 \\ 
49 & 77449 & 77771 &  & 77641 &  & 77833 &  & 77449 & 78176.6 &  & 77449 & 77794.2 &  & 77449 & 78091.2  &  & 77449 & 77449 & 77449 & 15.4 \\ 
50 & 31092 & 31810 &  & 31565 &  & 31292 &  & 31092 & 31580.6 &  & 31092 & 31359.8 &  & 31092 & 31194.6  &  & 31092 & 31092 & 31092 & 17.0 \\ 
51 & 49208 & 49907 &  & 49927 &  & 49761 &  & 49208 & 50082.5 &  & 49208 & 49638.3 &  & 49208 & 49510.6  &  & 49208 & 49373.2 & 49621 & 8.2 \\ 
52 & 93045 & 94175 &  & 94603 &  & 93106 &  & 93045 & 94653.3 &  & 93045 & 93722.2 &  & 93045 & 93782.8  &  & 93045 & 93045 & 93045 & 15.5 \\ 
53 & 84841 & 86891 &  & 84841 &  & 84841 &  & 84841 & 86465.3 &  & 84841 & 85422.4 &  & 84841 & 86566.7  &  & 84841 & 84841 & 84841 & 13.5 \\ 
54 & 118809 & 118809 &  & 119226 &  & 119074 &  & 118809 & 120150.8 &  & 118809 & 119194.9 &  & 118809 & 119639.6 &  & 118809 & 118872 & 119117 & 11.7 \\ 
55 & 64315 & 68649 &  & 66006 &  & 65400 &  & 64315 & 66055.4 &  & 64315 & 65240.2 &  & 64315 & 65106.7 &  & 64315 & 64315 & 64315 & 12.8 \\ 
56 & 74889 & 75490 &  & 75367 &  & 74940 &  & 74889 & 75472.9 &  & 74889 & 75038.9 &  & 74889 & 75060.2 &  & 74889 & 74894.1 & 74940 & 13.8 \\ 
57 & 63514 & 64575 &  & 64552 &  & 64575 &  & 63514 & 65195.4 &  & 63514 & 64195.4 &  & 63514 & 64494.3 &  & 63514 & 63593.2 & 64306 & 9.5 \\ 
58 & 45322 & 45680 &  & 45322 &  & 45322 &  & 45322 & 46286.1 &  & 45322 & 45495.2 &  & 45322 & 45781.0 &  & 45322 & 45322 & 45322 & 11.1 \\ 
59 & 50999 & 52001 &  & 52207 &  & 51649 &  & 50999 & 51954.0 &  & 50999 & 51463.4 &  & 50999 & 51348.2 &  & 50999 & 51162.8 & 51233 & 10.9 \\ 
60 & 60765 & 63342 &  & 60765 &  & 61755 &  & 60765 & 62498.8 &  & 60765 & 61843.8 &  & 60765 & 61397.4 &  & 60765 & 60866.7 & 61782 & 7.7 \\ 
61 & 75916 & 75916 &  & 75916 &  & 75916 &  & 75916 & 75998.5 &  & 75916 & 75916 &  & 75916 & 76094.9 &  & 75916 & 75916 & 75916 & 28.8 \\ 
62 & 44769 & 44769 &  & 44769 &  & 44769 &  & 44769 & 44840.0 &  & 44769 & 44775 &  & 44769 & 44833.1 &  & 44769 & 44769 & 44769 & 24.7 \\ 
63 & 75317 & 75317 &  & 75317 &  & 75317 &  & 75317 & 75536.9 &  & 75317 & 75317 &  & 75317 & 75627.0 &  & 75317 & 75317 & 75317 & 27.1 \\ 
64 & 92572 & 92572 &  & 92572 &  & 92572 &  & 92572 & 92609.8 &  & 92572 & 92572 &  & 92572 & 92650.5 &  & 92572 & 92572 & 92572 & 23.1 \\ 
65 & 126696 & 126696 &  & 126696 &  & 126696 &  & 126696 & 127080.1 &  & 126696 & 126696 &  & 126696 & 127428.5 &  & 126696 & 126696 & 126696 &  27.5 \\ 
66 & 59685 & 59685 &  & 59685 &  & 59685 &  & 59685 & 59927.3 &  & 59685 & 59685 &  & 59685 & 60034.8 &  & 59685 & 59685 & 59685 & 32.0 \\ 
67 & 29390 & 29390 &  & 29390 &  & 29390 &  & 29390 & 29415.8 &  & 29390 & 29390.4 &  & 29390 & 29397.0 &  & 29390 & 29390 & 29390 & 21.9 \\ 
68 & 22120 & 22120 &  & 22120 &  & 22120 &  & 22120 & 22264.6 &  & 22120 & 22159.2 &  & 22120 & 22143.0 &  & 22120 & 22120 & 22120 & 26.3 \\ 
69 & 71118 & 71118 &  & 71118 &  & 71118 &  & 71118 & 71307.7 &  & 71118 & 71118 &  & 71118 & 71164.6 &  & 71118 & 71118 & 71118 & 23.8 \\ 
70 & 75102 & 75102 &  & 75102 &  & 75102 &  & 75102 & 75427.4 &  & 75102 & 75123 &  & 75102 & 75433.9 &  & 75102 & 75102 & 75102 & 26.5 \\ 
71 & 145007 & 145771 &  & 145007 &  & 145007 &  & 145007 & 147119.7 &  & 145007 & 145561.2 &  & 145007 & 146653.5 &  & 145007 & 145272 & 146121 & 21.3 \\ 
72 & 43286 & 43994 &  & 43904 &  & 43286 &  & 43286 & 45678.3 &  & 43286 & 43706.4 &  & 43286 & 44177.7 &  & 43286 & 43286 & 43286 & 18.7 \\ 
73 & 28785 & 28785 &  & 28785 &  & 28785 &  & 28785 & 29054.4 &  & 28785 & 28803.8 &  & 28785 & 29129.3 &  & 28785 & 28785 & 28785 & 18.1 \\ 
74 & 29777 & 30734 &  & 30313 &  & 30136 &  & 29777 & 30765.0 &  & 29777 & 30248 &  & 29777 & 30378.2 &  & 29777 & 29777 & 29777 & 15.0 \\ 
75 & 21602 & 21602 &  & 21602 &  & 21602 &  & 21602 & 22450.0 &  & 21602 & 21790.8 &  & 21602 & 22130.9 &  & 21602 & 21617.6 & 21758 & 15.2 \\ 
76 & 53555 & 53899 &  & 53555 &  & 54024 &  & 53555 & 54529.2 &  & 53555 & 53752.2 &  & 53555 & 53819.2 &  & 53555 & 53717.6 & 54024 & 13.9 \\ 
77 & 31817 & 31937 &  & 32237 &  & 31817 &  & 31937 & 33374.4 &  & 31817 & 32126.4 &  & 31817 & 32797.0 &  & 31817 & 31973.6 & 32237 & 21.2 \\ 
78 & 19462 & 19660 &  & 19462 &  & 19462 &  & 19462 & 20381.8 &  & 19462 & 19481.8 &  & 19462 & 19871.3 &  & 19462 & 19481.8 & 19660 & 11.2 \\ 
79 & 114999 & 114999 &  & 114999 &  & 114999 &  & 114999 & 116196.8 &  & 114999 & 114999 &  & 114999 & 116054.9 &  & 114999 & 114999 & 114999 & 14.5 \\ 
80 & 18157 & 18157 &  & 18157 &  & 18157 &  & 18157 & 19274.4 &  & 18157 & 18300.7 &  & 18157 & 18660.2 &  & 18157 & 18157 & 18157 & 13.6 \\ 
81 & 383485 & 383703 &  & 383485 &  & 383485 &  & 383485 & 383903.1 &  & 383485 & 383485 &  & 383485 & 383662.7 &  & 383485 & 383485 & 383485 & 13.3 \\ 
82 & 409479 & 409544 &  & 409544 &  & 409479 &  & 409479 & 409815.6 &  & 409479 & 409544 &  & 409479 & 409922.4 &  & 409479 & 409486 & 409544 & 20.5 \\ 
83 & 458752 & 458787 &  & 458752 &  & 458752 &  & 458752 & 458922.1 &  & 458752 & 458771.8 &  & 458752 & 458980.5 &  & 458752 & 458756 & 458787 & 23.1 \\ 
84 & 329670 & 329670 &  & 329670 &  & 329670 &  & 329670 & 329969.7 &  & 329670 & 329800.5 &  & 329670 & 329978.1 &  & 329670 & 329675 & 329720 & 19.2 \\ 
85 & 554766 & 555130 &  & 554993 &  & 554766 &  & 554766 & 555107.2 &  & 554766 & 554822 &  & 554766 & 555118.6 &  & 554766 & 554771 & 554773 & 25.6 \\ 
86 & 361417 & 361417 &  & 361417 &  & 361417 &  & 361417 & 361826.3 &  & 361417 & 361417 &  & 361417 & 361576.4 &  & 361417 & 361417 & 361417 & 20.1 \\ 
87 & 398551 & 398551 &  & 398670 &  & 398551 &  & 398551 & 398623.8 &  & 398551 & 398562.9 &  & 398551 & 398645.1 &  & 398551 & 398551 & 398551 & 17.3 \\ 
88 & 433186 & 433519 &  & 433186 &  & 433244 &  & 433186 & 433564.4 &  & 433186 & 433235.6 &  & 433186 & 433463.8 &  & 433186 & 433219 & 433244 & 22.7 \\ 
89 & 410092 & 410092 &  & 410092 &  & 410092 &  & 410092 & 410187.3 &  & 410092 & 410092 &  & 410092 & 410411.9 &  & 410092 & 410092 & 410092 & 18.1 \\ 
90 & 401653 & 401653 &  & 401653 &  & 401653 &  & 401653 & 401825.5 &  & 401653 & 401663.2 &  & 401653 & 401843.7 &  & 401653 & 401653 & 401653  & 17.4 \\ 
91 & 339933 & 343029 &  & 340508 &  & 339933 &  & 339933 & 340079.3 &  & 339933 & 339961.8 &  & 339933 & 340025.4 &  & 339933 & 339933 & 339933  & 27.0 \\ 
92 & 361152 & 361152 &  & 361152 &  & 361152 &  & 361152 & 362030.6 &  & 361152 & 361399.5 &  & 361152 & 362196.9 &  & 361152 & 361152 & 361152  & 8.6 \\ 
93 & 403423 & 406728 &  & 404548 &  & 404917 &  & 403423 & 405344.0 &  & 403423 & 404407.6 &  & 403423 & 404958.1 &  & 403423 & 403456 & 403757  & 11.5 \\ 
94 & 332941 & 332983 &  & 333020 &  & 332949 &  & 332941 & 333319.3 &  & 332949 & 332979.7 &  & 332941 & 333178.9 &  & 332941 & 332968 & 333020  & 7.6 \\ 
95 & 516926 & 521208 &  & 517011 &  & 517646 &  & 516926 & 518751.3 &  & 516926 & 517316.5 &  & 516926 & 518849.9 &  & 516926 & 517115 & 517398  & 11.0 \\ 
96 & 455448 & 459321 &  & 457631 &  & 457631 &  & 455448 & 457814.4 &  & 455448 & 456178.1 &  & 455448 & 458241.3 &  & 455448 & 455650 & 457249  & 12.0 \\ 
97 & 407590 & 410889 &  & 409263 &  & 407590 &  & 407590 & 408437.8 &  & 407590 & 407590 &  & 407590 & 408981.6 &  & 407590 & 407590 & 407590 &  12.0 \\ 
98 & 520582 & 522630 &  & 523486 &  & 520582 &  & 520582 & 522055.2 &  & 520582 & 521145.8 &  & 520582 & 522093.9 &  & 520582 & 520582 & 520582  & 14.1 \\ 
99 & 363518 & 365149 &  & 364442 &  & 363977 &  & 363518 & 364803.2 &  & 363518 & 364000.7 &  & 363518 & 364709.5 &  & 363518 & 363728 & 363977  & 10.7 \\ 
100 & 431736 & 432714 &  & 431736 &  & 432068 &  & 431736 & 433105.8 &  & 432068 & 432396 &  & 431736 & 432781.5 &  & 431736 & 431736 & 431736  & 10.8 \\ 
101 & 352990 & 352990 &  & 352990 &  & 352990 &  & 352990 & 353033.1 &  & 352990 & 352990 &  & 352990 & 353004.9 &  & 352990 & 352990 & 352990  & 17.8 \\ 
102 & 492572 & 493069 &  & 492748 &  & 492572 &  & 492572 & 492832.6 &  & 492572 & 492572 &  & 492572 & 492914.3 &  & 492572 & 492572 & 492572  & 14.2 \\ 
103 & 378602 & 378602 &  & 378602 &  & 378602 &  & 378602 & 378834.0 &  & 378602 & 378602 &  & 378602 & 378934.1 &  & 378602 & 378602 & 378602  & 15.0 \\ 
104 & 357963 & 357963 &  & 357963 &  & 357963 &  & 357963 & 358174.6 &  & 357963 & 357995.6 &  & 357963 & 358077.2 &  & 357963 & 357968 & 358017 & 18.2 \\ 
105 & 450806 & 450806 &  & 450806 &  & 450806 &  & 450806 & 450812.4 &  & 450806 & 450806.2 &  & 450806 & 450889.9 &  & 450806 & 450806 & 450806 & 19.1 \\ 
106 & 454379 & 455152 &  & 454379 &  & 454379 &  & 454379 & 454851.6 &  & 454379 & 454379 &  & 454379 & 455091.8 &  & 454379 & 454379 & 454379 & 22.5 \\ 
107 & 352766 & 352867 &  & 352766 &  & 352766 &  & 352766 & 353002.3 &  & 352766 & 352826.2 &  & 352766 & 353153.6 &  & 352766 & 352766 & 352766 & 18.1 \\ 
108 & 460793 & 460793 &  & 460793 &  & 460793 &  & 460793 & 461112.6 &  & 460793 & 460793 &  & 460793 & 461390.3 &  & 460793 & 460793 & 460793 & 28.7 \\ 
109 & 413004 & 413004 &  & 413004 &  & 413004 &  & 413004 & 413426.1 &  & 413004 & 413130.2 &  & 413004 & 413643.8 &  & 413004 & 413006 & 413019 & 16.9 \\ 
110 & 418769 & 418769 &  & 418769 &  & 418769 &  & 418769 & 419030.4 &  & 418769 & 418834.2 &  & 418769 & 418954.5 &  & 418769 & 418769 & 418769 & 13.4 \\ 
111 & 342752 & 342752 &  & 342752 &  & 342752 &  & 342752 & 343768.0 &  & 342752 & 342805.2 &  & 342752 & 343687.6 &  & 342752 & 342752 & 342752 & 17.4 \\ 
112 & 367110 & 369237 &  & 367110 &  & 367110 &  & 367110 & 369592.0 &  & 367110 & 367970.4 &  & 367110 & 368861.7 &  & 367110 & 367110 & 367110 & 17.6 \\ 
113 & 259649 & 260176 &  & 260872 &  & 259649 &  & 259649 & 259909.2 &  & 259649 & 259676.9 &  & 259649 & 260246.4 &  & 259649 & 259649 & 259649 & 9.6 \\ 
114 & 463474 & 464136 &  & 465503 &  & 463474 &  & 463474 & 465440.2 &  & 463474 & 464377.1 &  & 463474 & 465440.2 &  & 463474 & 463508 & 463813 & 8.4 \\ 
115 & 456890 & 457874 &  & 457289 &  & 457189 &  & 456890 & 458414.7 &  & 456890 & 457046.1 &  & 456890 & 458414.7 &  & 456890 & 456923 & 457189 & 17.0 \\ 
116 & 530601 & 532456 &  & 530803 &  & 530601 &  & 530601 & 531410.8 &  & 530601 & 530614.4 &  & 530601 & 531410.8 &  & 530601 & 530601 & 530601 & 10.9 \\ 
117 & 502840 & 503199 &  & 502840 &  & 503046 &  & 502840 & 503600.9 &  & 502840 & 502991.5 &  & 502840 & 503600.9 &  & 502840 & 502840 & 502840 & 13.8 \\ 
118 & 349749 & 350729 &  & 349749 &  & 349749 &  & 349749 & 352050.9 &  & 349749 & 350326.2 &  & 349749 & 352050.9 &  & 349749 & 349749 & 349749 & 15.5 \\ 
119 & 573046 & 573046 &  & 573046 &  & 573046 &  & 573046 & 573759.3 &  & 573046 & 573122.1 &  & 573046 & 573759.3 &  & 573046 & 573046 & 573046 & 12.0 \\ 
120 & 396183 & 396183 &  & 396183 &  & 396183 &  & 396183 & 398005.1 &  & 396183 & 396592.5 &  & 396183 & 398005.1 &  & 396183 & 396183 & 396183 & 15.4 \\   
\hline
&  &  &  &  &  &  &  &  &  &  &  &  &  &  &  &  &  &  & Mean & 13.07 
\label{Comparison}
\end{longtable}
}
\end{center}

\vspace{-1cm}

Table \ref{SummaryComparison} presents a summary of the results found by ILS-RVND$_{Fast}$ compared with those achieved by several heuristics from the literature. A direct comparison with some of the algorithms such as ACO$_{\text{AP}}$, DPSO, DDE and GVNS becomes quite hard because the average results were not reported by the authors. With respect to the best solutions, our algorithm clearly outperforms these four by a good margin. Even our average and worst solutions are, in most cases, better or equal than the best ones of such methods. 

The average and worst solutions of ILS-RVND$_{Fast}$ are always equal or better than the average solutions of ILS$_{\text{XLC}}$. The same happened to ILS-RVND$_{\text{SBP}}$ and LOX$\oplus$B, except for few cases where the worst solutions of ILS-RVND$_{Fast}$ were not better than the average solution of these two methods. Worst solutions were only reported for ILS-RVND$_{\text{SBP}}$, where all them are either equal or worse than those of ILS-RVND$_{Fast}$. The results suggest that our algorithm is very competitive in terms of solution quality. 


\begin{table}[!htbp]
\caption{Summary of the results found by ILS-RVND$_{Fast}$ compared to several heuristic methods from the literature}
\centering
\scriptsize
\onehalfspacing
\setlength{\tabcolsep}{1.5mm}
\label{SummaryComparison}
\begin{tabular}{lccccccc}
\hline
 & ACO$_{\text{AP}}$ & DPSO & DDE & GVNS & ILS$_{\text{XLC}}$ &  ILS-RVND$_{\text{SBP}}$ & LOX$\oplus$B \\ \hline
$\#Best$ improved & 84 & 66 & 52 & 44 & 6 & 20 & 1 \\ 
$\#Best$ equaled & 36 & 54 & 68 & 76 & 114 & 100 & 118 \\
$\#Best$ worse & 0 & 0 & 0 & 0 & 0 & 0 & 1 \\ 
$\#Avg.$ better than the Best & 84 & 65 & 51 & 42 & 2 & 19 & 0 \\ 
$\#Avg.$ equal to the Best & 34 & 49 & 57 & 64 & 76 & 74 & 76 \\ 
$\#Avg.$ improved & -- & -- & -- & -- & 101 & 83 & 101 \\ 
$\#Avg.$ equaled & -- & -- & -- & -- & 19 & 37 & 19 \\
$\#Avg.$ worse & -- & -- & -- & -- & 0 & 0 & 0 \\ 
$\#Worst$ better than the Best & 82 & 59 & 47 & 34 & 0 & 13 & 0 \\ 
$\#Worst$ equal to the Best & 34 & 52 & 51 & 69 & 76 & 78 & 76 \\ 
$\#Worst$ better than the Avg.& -- & -- & -- & -- & 101 & 69 & 91 \\ 
$\#Worst$ equal to the Avg. & -- & -- & -- & -- & 19 & 37 & 20 \\ 
$\#Worst$ improved & -- & -- & -- & -- & -- & 82 & -- \\ 
$\#Worst$ equaled & -- & -- & -- & -- & -- & 38 & -- \\ 
$\#Worst$ worse & -- & -- & -- & -- & -- & 0 & -- \\ \hline

\end{tabular}
\end{table}

Table \ref{Target} shows the time in seconds spent by ILS-RVND$_{Fast}$ to find or improve the solutions found by the algorithms that were chosen for comparison. In this case the stopping criterion no longer depends on the number of restarts, but on a target value from the literature. The algorithm was executed 10 times for each instance and we report the mean (arithmetic), median, minimum and maximum of the average times by group required to find or improve the target value. We also report the average values considering all instances at once (last column of the table).

\begin{table}[!htbp]
\caption{Average time in seconds required by ILS-RVND$_{Fast}$ to find or improve the solutions found by different algorithms from the literature}
\scriptsize
\centering
\setlength{\tabcolsep}{1.7mm}
\label{Target}
\begin{tabular}{cc|lrrrrrrrrrrrrr}
\hline
\multicolumn{1}{c}{} & \multicolumn{1}{c}{} &  & \multicolumn{ 12}{c}{Group} \\ \cline{4-15}\noalign{\smallskip}
\multicolumn{1}{c}{} & \multicolumn{1}{c}{} &  & \multicolumn{1}{c}{1} & \multicolumn{1}{c}{2} & \multicolumn{1}{c}{3} & \multicolumn{1}{c}{4} & \multicolumn{1}{c}{5} & \multicolumn{1}{c}{6} & \multicolumn{1}{c}{7} & \multicolumn{1}{c}{8} & \multicolumn{1}{c}{9} & \multicolumn{1}{c}{10} & \multicolumn{1}{c}{11} & \multicolumn{1}{c}{12} & \multicolumn{1}{c}{\up{All}} \\ 
\hline\noalign{\smallskip}
\multicolumn{1}{c|}{\multirow{ 4}{*}{\rotatebox{90}{ACO$_{\text{AP}}$}}} & \multirow{ 4}{*}{\rotatebox{90}{Best}} & Mean & 0.5 & 0.2 & 0.2 & 0.2 & 1.2 & 1.0 & 2.9 & 3.4 & 7.2 & 2.2 & 2.9 & 1.7 & 2.0 \\ 
\multicolumn{ 1}{c|}{} & \multicolumn{ 1}{c|}{} & Median & 0.3 & 0.1 & $< 0.1$ & $< 0.1$ & 0.7 & 0.9 & 2.7 & 1.8 & 1.9 & 1.1 & 2.3 & 1.5 & 0.9 \\ 
\multicolumn{ 1}{c|}{} & \multicolumn{ 1}{c|}{} & Min & 0.3 & $< 0.1$ & $< 0.1$ & $< 0.1$ & 0.2 & 0.5 & 0.4 & 0.7 & 0.4 & 0.5 & 0.3 & 0.3 & $< 0.1$\\ 
\multicolumn{ 1}{c|}{} & \multicolumn{ 1}{c|}{} & Max & 1.6 & 0.7 & 1.2 & 1.7 & 3.5 & 1.5 & 5.5 & 12.3 & 47.2 & 11.6 & 6.7 & 4.5 & 47.2\\ 
\multicolumn{1}{c}{} & \multicolumn{1}{c}{} &  &  &  &  &  &  &  &  &  &  &  &  &  \\  
\multicolumn{1}{c|}{\multirow{ 4}{*}{\rotatebox{90}{DPSO}}} & \multirow{ 4}{*}{\rotatebox{90}{Best}} & Mean & 0.6 & 0.3 & 0.4 & 0.4 & 3.1 & 2.8 & 5.1 & 7.5 & 3.8 & 2.2 & 3.7 & 3.3 & 2.8\\  
\multicolumn{ 1}{c|}{} & \multicolumn{ 1}{c|}{} & Median & 0.4 & 0.3 & $< 0.1$ & $< 0.1$ & 2.8 & 2.1 & 3.9 & 6.7 & 2.9 & 1.8 & 2.9 & 3.3 & 2.2\\ 
\multicolumn{ 1}{c|}{} & \multicolumn{ 1}{c|}{} & Min & 0.2 & $< 0.1$ & $< 0.1$ & $< 0.1$ & 0.9 & 0.4 & 2.2 & 3.3 & 1.9 & 0.3 & 0.7 & 0.1 & $< 0.1$\\ 
\multicolumn{ 1}{c|}{} & \multicolumn{ 1}{c|}{} & Max & 1.3 & 1.1 & 2.3 & 3.7 & 9.6 & 8.7 & 15.3 & 15.5 & 8.1 & 6.3 & 7.9 & 6.0 & 15.5\\ 
 \multicolumn{1}{c}{} & \multicolumn{1}{c}{} &  &  &  &  &  &  &  &  &  &  &  &  &  \\ 
\multicolumn{1}{c|}{\multirow{ 4}{*}{\rotatebox{90}{DDE}}} & \multirow{ 4}{*}{\rotatebox{90}{Best}} & Mean & 2.1 & 0.7 & 1.0 & 0.5 & 4.9 & 3.9 & 5.5 & 9.0 & 12.5 & 5.2 & 5.8 & 4.5 & 4.6 \\  
\multicolumn{ 1}{c|}{} & \multicolumn{ 1}{c|}{} & Median & 1.6 & 0.4 & $< 0.1$ & $< 0.1$ & 3.8 & 4.0 & 3.1 & 8.0 & 4.6 & 3.3 & 5.4 & 4.4 & 3.0 \\ 
\multicolumn{ 1}{c|}{} & \multicolumn{ 1}{c|}{} & Min & 1.3 & $< 0.1$ & $< 0.1$ & $< 0.1$ & 0.9 & 1.6 & 0.8 & 3.4 & 1.0 & 1.0 & 1.6 & 0.9 & $< 0.1$\\ 
\multicolumn{ 1}{c|}{} & \multicolumn{ 1}{c|}{} & Max & 7.3 & 2.7 & 7.9 & 5.1 & 15.5 & 6.9 & 21.3 & 22.3 & 81.8 & 23.9 & 12.9 & 8.7 & 81.8\\ 
\multicolumn{1}{c}{} & \multicolumn{1}{c}{} &  &  &  &  &  &  &  &  &  &  &  &  &  \\  
\multicolumn{1}{c|}{\multirow{ 4}{*}{\rotatebox{90}{GVNS}}} & \multirow{ 4}{*}{\rotatebox{90}{Best}} & Mean & 2.3 & 1.4 & 11.4 & 1.0 & 3.6 & 5.2 & 5.8 & 15.0 & 14.9 & 4.0 & 4.7 & 4.8 & 6.2\\  
\multicolumn{ 1}{c|}{} & \multicolumn{ 1}{c|}{} & Median & 2.2 & 1.4 & $< 0.1$ & $< 0.1$ & 2.6 & 5.3 & 4.1 & 6.3 & 5.8 & 3.5 & 3.5 & 3.8 & 3.1 \\ 
\multicolumn{ 1}{c|}{} & \multicolumn{ 1}{c|}{} & Min & 0.6 & $< 0.1$ & $< 0.1$ & $< 0.1$ & 0.8 & 2.8 & 1.6 & 3.0 & 2.8 & 2.6 & 1.9 & 2.1 & $< 0.1$ \\ 
\multicolumn{ 1}{c|}{} & \multicolumn{ 1}{c|}{} & Max & 4.4 & 3.0 & 113.1 & 9.9 & 8.1 & 8.1 & 15.4 & 82.9 & 81.3 & 10.1 & 11.9 & 8.6 & 113.1 \\ 
\multicolumn{1}{c}{} & \multicolumn{1}{c}{} &  &  &  &  &  &  &  &  &  &  &  &  &  \\ 
\multicolumn{1}{c|}{\multirow{ 9}{*}{\rotatebox{90}{ILS$_{\text{XLC}}$}}} & \multirow{ 4}{*}{\rotatebox{90}{Best}} & Mean & 21.4 & 17.0 & 282.2$^1$ & 5.5 & 6.0 & 8.3 & 3.8 & 9.2 & 14.8 & 11.1 & 4.2 & 8.3 & 32.6$^2$ \\ 
\multicolumn{ 1}{c|}{} & \multicolumn{ 1}{c|}{} & Median & 17.8 & 8.1 & $< 0.1$ & $< 0.1$ & 5.2 & 8.0 & 3.0 & 7.5 & 6.5 & 5.6 & 3.9 & 4.2 & 4.9 \\ 
\multicolumn{ 1}{c|}{} & \multicolumn{ 1}{c|}{} & Min & 2.3 & $< 0.1$ & $< 0.1$ & $< 0.1$ & 1.6 & 4.2 & 1.8 & 2.9 & 1.0 & 2.0 & 1.7 & 2.2 & $< 0.1$ \\ 
\multicolumn{ 1}{c|}{} & \multicolumn{ 1}{c|}{} & Max & 54.6 & 47.9 & 2820.4 & 54.8 & 13.4 & 13.7 & 8.8 & 17.0 & 52.1 & 32.6 & 8.6 & 42.0 & 2820.4 \\ 
\multicolumn{1}{c|}{} & \multicolumn{1}{c}{} &  &  &  &  &  &  &  &  &  &  &  &  &  \\ 
\multicolumn{ 1}{c|}{} & \multirow{ 4}{*}{\rotatebox{90}{Avg}} & Mean & 1.5 & 1.6 & 0.3 & 0.4 & 2.4 & 2.1 & 2.5 & 2.6 & 2.8 & 2.6 & 1.9 & 2.5 & 1.9 \\ 
\multicolumn{ 1}{c|}{} & \multicolumn{ 1}{c|}{} & Median & 1.3 & 1.8 & $< 0.1$ & $< 0.1$ & 1.8 & 2.2 & 2.1 & 2.6 & 2.4 & 2.7 & 1.6 & 2.3 & 1.9 \\ 
\multicolumn{ 1}{c|}{} & \multicolumn{ 1}{c|}{} & Min & 0.9 & $< 0.1$ & $< 0.1$ & $< 0.1$ & 1.3 & 0.9 & 1.7 & 1.0 & 1.6 & 1.3 & 0.7 & 0.9 & $< 0.1$ \\ 
\multicolumn{ 1}{c|}{} & \multicolumn{ 1}{c|}{} & Max & 3.0 & 2.6 & 2.4 & 4.4 & 4.1 & 3.1 & 4.1 & 5.2 & 5.3 & 4.8 & 4.5 & 4.0 & 5.3 \\ 
\multicolumn{1}{c}{} & \multicolumn{1}{c}{} &  &  &  &  &  &  &  &  &  &  &  &  &  \\ 
\multicolumn{1}{c|}{\multirow{ 9}{*}{\rotatebox{90}{ILS-RVND$_{\text{SBP}}$}}} & \multirow{ 4}{*}{\rotatebox{90}{Best}} & Mean & 4.1 & 2.6 & 5.3 & 4.3 & 5.5 & 11.4 & 4.6 & 17.1 & 18.4 & 9.3 & 5.2 & 7.1 & 7.9 \\ 
\multicolumn{ 1}{c|}{} & \multicolumn{ 1}{c|}{} & Median & 3.2 & 2.9 & $< 0.1$ & $< 0.1$ & 5.6 & 10.5 & 3.0 & 8.3 & 5.0 & 4.9 & 4.6 & 4.7 & 4.1 \\ 
\multicolumn{ 1}{c|}{} & \multicolumn{ 1}{c|}{} & Min & 1.7 & $< 0.1$ & $< 0.1$ & $< 0.1$ & 2.2 & 5.7 & 1.4 & 3.4 & 1.6 & 2.9 & 1.8 & 3.0 & $< 0.1$ \\ 
\multicolumn{ 1}{c|}{} & \multicolumn{ 1}{c|}{} & Max & 7.3 & 5.2 & 52.0 & 43.2 & 8.7 & 25.6 & 19.3 & 88.6 & 67.7 & 35.7 & 14.4 & 29.8 & 88.6 \\ 
\multicolumn{1}{c|}{} & \multicolumn{1}{c}{} &  &  &  &  &  &  &  &  &  &  &  &  &  \\ 
\multicolumn{ 1}{c|}{} & \multirow{ 4}{*}{\rotatebox{90}{Avg}} & Mean & 1.9 & 1.6 & 1.5 & 0.6 & 4.1 & 5.0 & 4.2 & 8.0 & 8.5 & 5.7 & 3.7 & 4.0 & 4.1 \\
\multicolumn{ 1}{c|}{} & \multicolumn{ 1}{c|}{} & Median & 2.0 & 1.8 & $< 0.1$ & $< 0.1$ & 3.4 & 5.3 & 4.2 & 7.3 & 4.3 & 3.7 & 3.8 & 3.8 & 3.2 \\ 
\multicolumn{ 1}{c|}{} & \multicolumn{ 1}{c|}{} & Min & 1.3 & $< 0.1$ & $< 0.1$ & $< 0.1$ & 2.0 & 2.6 & 1.2 & 5.0 & 1.4 & 2.3 & 1.4 & 0.6 & $< 0.1$ \\ 
\multicolumn{ 1}{c|}{} & \multicolumn{ 1}{c|}{} & Max & 2.5 & 3.1 & 14.1 & 5.5 & 9.4 & 6.5 & 8.1 & 16.7 & 41.8 & 24.4 & 7.6 & 9.6 & 41.8 \\ 
\multicolumn{1}{c}{} & \multicolumn{1}{c}{} &  &  &  &  &  &  &  &  &  &  &  &  &  \\ 
\multicolumn{1}{c|}{\multirow{ 9}{*}{\rotatebox{90}{LOX$\oplus$B}}} & \multirow{ 4}{*}{\rotatebox{90}{Best}} & Mean & 48.1 & 87.6 & 1685.8$^3$ & 5.5 & 6.0 & 8.3 & 3.8 & 15.8 & 14.8 & 11.1 & 4.2 & 8.3 & 158.3$^4$ \\ 
\multicolumn{ 1}{c|}{} & \multicolumn{ 1}{c|}{} & Median & 21.0 & 28.9 & $< 0.1$ & $< 0.1$ & 5.2 & 8.0 & 3.0 & 7.5 & 6.5 & 5.6 & 3.9 & 4.2 & 4.9 \\ 
\multicolumn{ 1}{c|}{} & \multicolumn{ 1}{c|}{} & Min & 2.3 & $< 0.1$ & $< 0.1$ & $< 0.1$ & 1.6 & 4.2 & 1.8 & 2.9 & 1.0 & 1.9 & 1.7 & 2.2 & $< 0.1$ \\ 
\multicolumn{ 1}{c|}{} & \multicolumn{ 1}{c|}{} & Max & 281.3 & 318.5 & 16856.4 & 54.8 & 13.4 & 13.7 & 8.8 & 82.9 & 52.1 & 32.6 & 8.6 & 42.0 & 16856.4 \\ 
\multicolumn{1}{c|}{} & \multicolumn{1}{c}{} &  &  &  &  &  &  &  &  &  &  &  &  &  \\ 
\multicolumn{ 1}{c|}{} & \multirow{ 4}{*}{\rotatebox{90}{Avg}} & Mean & 4.0 & 3.3 & 2.2 & 0.6 & 3.9 & 4.7 & 2.4 & 5.6 & 2.6 & 2.2 & 1.7 & 2.6 & 3.0 \\ 
\multicolumn{ 1}{c|}{} & \multicolumn{ 1}{c|}{} & Median & 3.5 & 3.5 & $< 0.1$ & $< 0.1$ & 3.4 & 4.8 & 2.5 & 4.7 & 2.8 & 2.2 & 1.5 & 1.8 & 2.4 \\ 
\multicolumn{ 1}{c|}{} & \multicolumn{ 1}{c|}{} & Min & 2.8 & $< 0.1$ & $< 0.1$ & $< 0.1$ & 1.8 & 1.4 & 0.9 & 1.9 & 0.5 & 1.5 & 0.5 & 0.5 & $< 0.1$ \\ 
\multicolumn{ 1}{c|}{} & \multicolumn{ 1}{c|}{} & Max & 7.6 & 6.6 & 20.0 & 5.6 & 7.9 & 10.3 & 4.7 & 11.8 & 4.5 & 3.8 & 4.2 & 8.4 & 20.0 \\ 
\multicolumn{1}{c}{} & \multicolumn{1}{c}{} &  &  &  &  &  &  &  &  &  &  &  &  &  \\ 
\multicolumn{ 2}{c}{\multirow{4}{*}{\rotatebox{90}{Optimal}}} & \multicolumn{ 1}{|l}{Mean} & 48.1 & 122.9 & 1685.8$^3$ & 5.5 & 6.0 & 8.3 & 3.8 & 15.8 & 14.8 & 11.1 & 4.2 & 8.3 & 161.2$^5$\\ 
\multicolumn{ 2}{l}{} & \multicolumn{ 1}{|l}{Median} & 21.0 & 28.9 & 0.01 & $< 0.1$ & 5.2 & 8.0 & 3.0 & 7.5 & 6.5 & 5.6 & 3.9 & 4.2 & 4.9 \\ 
\multicolumn{ 2}{l}{} & \multicolumn{ 1}{|l}{Min} & 2.3 & $< 0.1$ & $< 0.1$ & $< 0.1$ & 1.6 & 4.2 & 1.8 & 2.9 & 1.0 & 2.0 & 1.7 & 2.2 & $< 0.1$ \\ 
\multicolumn{ 2}{l}{} & \multicolumn{ 1}{|l}{Max} & 281.3 & 605.0 & 16856.4 & 54.8 & 13.4 & 13.7 & 8.8 & 82.9 & 52.1 & 32.6 & 8.6 & 42.0 & 16856.4 \\ \hline\noalign{\smallskip}
\multicolumn{11}{l}{\tiny $^1$: 0.1 seconds disregarding instance 24} \\
\multicolumn{11}{l}{\tiny $^2$: 9.2 seconds disregarding instance 24} \\
\multicolumn{11}{l}{\tiny $^3$: 0.2 seconds disregarding instance 24} \\
\multicolumn{11}{l}{\tiny $^4$: 18.0 seconds disregarding instance 24} \\
\multicolumn{11}{l}{\tiny $^5$: 20.9 seconds disregarding instance 24} \\
\end{tabular}
\end{table}

On one hand, it can be observed from Table \ref{Target} that ILS-RVND$_{Fast}$ was capable of finding or improving the best solutions of ACO$_{\text{AP}}$, DPSO, DDE, GVNS and ILS-RVND$_{\text{SBP}}$ in only 2.0, 2.8, 4.6, 6.2 and 7.9 seconds, on average, respectively. On the other hand, ILS-RVND$_{Fast}$ spent, on average, 32.6, 158.3 and 161.2 seconds to find or improve the best solutions of ILS$_{\text{XLC}}$, LOX$\oplus$B and the optimal solutions, respectively. However, if we disregard instance 24, these values significantly decrease to 9.2, 18.0 and 20.9 seconds, respectively. 

Moreover, ILS-RVND$_{Fast}$ found or improved the average solutions of ILS$_{\text{XLC}}$, ILS-RVND$_{\text{SBP}}$ and LOX$\oplus$B in only 1.9, 4.1 and 3.0 seconds, respectively, on average. It is worth mentioning that the average solutions of ILS$_{\text{XLC}}$ in \cite{Xu2013} and LOX$\oplus$B in \cite{Xu2014} were obtained in 100 seconds on an Intel Core i3 3.10 GHz with 2.0 GB of RAM, whereas the average solutions of ILS-RVND$_{\text{SBP}}$ were obtained in 23.4 seconds on an Intel Core i5 3.20 GHz with 4.0 GB of RAM. The hardware configuration of these two machines is slightly inferior than the one used in our experiments (Intel Core i7 with 3.40 GHz and 16 GB of RAM), and thus does not justify the considerable difference in terms of CPU time between ILS-RVND$_{Fast}$ and the three other methods. Therefore, from Tables \ref{Comparison}-\ref{Target}, we can conclude that ILS-RVND$_{Fast}$ is, on average, remarkably faster and clearly more efficient than those three heuristic algorithms, which are the 
best ones available in the literature for problem $1|s_{ij}|\sum w_jT_j$.

\subsection{Impact of the proposed local search limitation strategy}
\label{sec:impact}

In this section we investigate the effect of the proposed local search limitation strategy on the performance of the algorithm. We start by analyzing the average percentage of moves that were not evaluated per neighborhood in all groups of instances, as shown in Table \ref{MovesNotEvaluated}. It can be verified that the proportion of moves that were not considered for evaluation varied according to the characteristics of the group, ranging, on average, from 69.6\% (Group 7) to 98.8\% (Group 2). 

\begin{table}[!htbp]
\caption{Average percentage of moves that were not evaluated}
\centering
\scriptsize
\onehalfspacing
\begin{tabular}{lcccccccccccc}
\hline
\multicolumn{1}{l}{} & \multicolumn{12}{c}{Group} \\ \cline{2-13}
\multicolumn{1}{c}{\up{Neighborhoods}} & 1 & 2 & 3 & 4 & 5 & 6 & 7 & 8 & 9 & 10 & 11 & 12 \\ \hline
1-block insertion & 87.7 & 96.2 & 89.4 & 94.9 & 81.7 & 93.9 & 78.5 & 91.6 & 89.3 & 94.5 & 88.5 & 94.9 \\ 
2-block insertion & 92.7 & 96.9 & 90.9 & 96.7 & 82.2 & 93.7 & 75.2 & 90.0 & 84.6 & 93.1 & 83.4 & 93.9 \\ 
3-block insertion & 95.0 & 97.9 & 91.4 & 96.6 & 82.0 & 93.6 & 73.7 & 89.8 & 80.3 & 91.8 & 78.8 & 92.6 \\ 
4-block insertion & 96.4 & 99.0 & 86.3 & 96.4 & 85.5 & 94.9 & 74.6 & 90.9 & 80.1 & 91.8 & 78.1 & 92.3 \\ 
5-block insertion & 96.6 & 99.5 & 86.7 & 97.1 & 83.7 & 95.6 & 73.3 & 91.1 & 80.7 & 91.0 & 78.9 & 92.5 \\ 
6-block insertion & 96.8 & 99.6 & 79.0 & 95.5 & 85.8 & 95.0 & 75.2 & 91.1 & 81.9 & 90.7 & 79.5 & 91.8 \\ 
7-block insertion & 97.5 & 99.5 & 78.7 & 95.5 & 84.1 & 95.1 & 72.6 & 89.8 & 81.7 & 91.0 & 79.0 & 92.4 \\ 
8-block insertion & 97.5 & 99.5 & 79.1 & 96.5 & 83.4 & 93.9 & 69.3 & 89.0 & 78.9 & 90.4 & 75.7 & 92.0 \\ 
9-block insertion & 96.5 & 99.5 & 79.4 & 94.0 & 80.4 & 94.0 & 65.1 & 87.5 & 77.7 & 90.6 & 74.8 & 90.2 \\ 
10-block insertion & 96.7 & 99.6 & 83.4 & 94.7 & 79.0 & 92.9 & 63.9 & 85.7 & 75.2 & 88.9 & 71.2 & 89.5 \\ 
11-block insertion & 96.4 & 99.4 & 79.1 & 94.9 & 76.7 & 92.5 & 61.4 & 84.5 & 71.3 & 87.3 & 69.8 & 89.1 \\ 
12-block insertion & 96.2 & 99.5 & 71.5 & 92.1 & 74.7 & 91.4 & 58.3 & 83.9 & 69.4 & 86.8 & 66.5 & 87.6 \\ 
13-block insertion & 94.9 & 99.3 & 72.7 & 92.8 & 72.4 & 90.4 & 55.0 & 82.6 & 67.0 & 86.2 & 63.5 & 86.0 \\ 
Swap & 89.7 & 98.3 & 87.3 & 96.8 & 81.7 & 95.4 & 78.0 & 92.5 & 90.1 & 94.8 & 89.7 & 95.7 \\ \hline
\multicolumn{1}{c}{Mean} & 95.0 & 98.8 & 82.5 & 95.3 & 80.9 & 93.7 & 69.6 & 88.6 & 79.1 & 90.6 & 77.0 & 91.5 \\ \hline
\end{tabular}
\label{MovesNotEvaluated}
\end{table}

Since a very large number of moves were not considered for evaluation, we decided to conduct an experiment to verify the level of accuracy of the limitation strategy. Table \ref{LostMoves} shows, for each neighborhood and for each group, the average percentage of improving moves that were not evaluated, here denoted as \emph{lost improving moves}. We can observe that the average percentage of lost improving moves was relatively small in all cases (never more than 11\%), thus ratifying the effectiveness of the proposed limitation strategy. 

\begin{table}[!htbp]
 \caption{Average percentage of improving moves that were not evaluated}
\centering
\scriptsize
\onehalfspacing
\begin{tabular}{lcccccccccccc}
\hline
\multicolumn{1}{l}{} & \multicolumn{12}{c}{Group} \\ \cline{2-13}
\multicolumn{1}{c}{\up{Neighborhoods}} & 1 & 2 & 3 & 4 & 5 & 6 & 7 & 8 & 9 & 10 & 11 & 12 \\ \hline
1-block insertion & 10.9 & 8.2 & 1.8 & 2.3 & 9.4 & 10.0 & 10.0 & 9.6 & 9.6 & 9.7 & 9.7 & 9.6 \\ 
2-block insertion & 10.0 & 8.7 & 1.3 & 1.0 & 9.9 & 9.6 & 9.8 & 9.9 & 10.0 & 10.0 & 9.8 & 9.7 \\ 
3-block insertion & 9.9 & 8.8 & 2.8 & 2.9 & 9.3 & 9.9 & 9.9 & 9.6 & 9.7 & 10.4 & 9.6 & 9.7 \\ 
4-block insertion & 9.9 & 7.8 & 2.6 & 1.2 & 9.8 & 9.9 & 9.6 & 9.4 & 9.9 & 10.0 & 9.8 & 10.0 \\ 
5-block insertion & 9.8 & 7.8 & 1.3 & 1.2 & 9.9 & 9.7 & 9.7 & 9.8 & 10.2 & 10.1 & 9.7 & 10.0 \\ 
6-block insertion & 9.3 & 6.8 & 2.0 & 1.0 & 9.9 & 10.5 & 9.6 & 9.6 & 10.4 & 10.3 & 10.1 & 10.3 \\ 
7-block insertion & 9.6 & 7.7 & 1.3 & 2.0 & 9.1 & 10.0 & 9.8 & 9.8 & 9.8 & 10.1 & 9.8 & 9.7 \\ 
8-block insertion & 9.1 & 7.0 & 1.4 & 1.3 & 9.7 & 9.7 & 9.6 & 9.9 & 9.7 & 9.0 & 9.3 & 10.2 \\ 
9-block insertion & 9.5 & 7.3 & 1.7 & 1.3 & 10.1 & 9.6 & 9.7 & 9.9 & 10.5 & 10.0 & 9.4 & 10.3 \\ 
10-block insertion & 9.7 & 7.5 & 1.6 & 1.3 & 10.5 & 10.3 & 9.7 & 9.4 & 9.8 & 10.4 & 9.5 & 10.0 \\ 
11-block insertion & 9.4 & 7.7 & 1.5 & 2.2 & 9.7 & 10.2 & 9.4 & 9.8 & 9.6 & 10.0 & 10.5 & 9.6 \\ 
12-block insertion & 9.5 & 8.2 & 1.6 & 1.4 & 9.0 & 10.6 & 9.8 & 9.5 & 9.4 & 9.9 & 9.9 & 10.4 \\ 
13-block insertion & 9.7 & 7.7 & 1.7 & 1.3 & 10.2 & 10.5 & 10.2 & 10.0 & 9.4 & 10.3 & 9.8 & 10.1 \\ 
Swap & 9.8 & 9.7 & 2.3 & 1.1 & 9.2 & 9.6 & 9.2 & 9.9 & 9.5 & 9.5 & 9.5 & 10.0 \\ \hline
\multicolumn{1}{c}{Mean} & 9.7 & 7.9 & 1.8 & 1.5 & 9.7 & 10.0 & 9.7 & 9.7 & 9.8 & 10.0 & 9.8 & 10.0 \\ \hline
\end{tabular}
\label{LostMoves}
\end{table}

Tables \ref{Time} and \ref{TimeFast} present, for every group of instances, the average time in seconds spent by each neighborhood in a complete execution of ILS-RVND and ILS-RVND$_{Fast}$, respectively. The impact of the limitation strategy on the average time spent during the local search is clearly visible. It is possible to verify that a neighborhood search in ILS-RVND$_{Fast}$ is, on average, about 6 times faster than in ILS-RVND. Also, we can see that the level of CPU time reduction for each group is proportional to the number of moves that were not evaluated, as shown in Table \ref{MovesNotEvaluated}.


\begin{table}[!htbp]
\caption{Average time in seconds spent by each neighborhood in ILS-RVND}
\centering
\scriptsize
\onehalfspacing
\begin{tabular}{lcccccccccccc}
\hline
\multicolumn{1}{l}{} & \multicolumn{12}{c}{Group} \\ \cline{2-13}
\multicolumn{1}{c}{\up{Neighborhoods}} & 1 & 2 & 3 & 4 & 5 & 6 & 7 & 8 & 9 & 10 & 11 & 12 \\ \hline
1-block insertion & 6.3 & 5.3 & 0.9 & 1.0 & 6.5 & 7.6 & 8.0 & 9.4 & 6.7 & 6.9 & 5.6 & 7.6 \\ 
2-block insertion & 6.2 & 5.1 & 0.8 & 0.9 & 5.9 & 7.0 & 6.9 & 8.0 & 6.1 & 6.3 & 5.2 & 7.0 \\ 
3-block insertion & 5.8 & 4.8 & 0.8 & 0.8 & 5.5 & 6.4 & 6.0 & 6.9 & 5.8 & 5.9 & 4.8 & 6.5 \\ 
4-block insertion & 5.8 & 4.8 & 0.7 & 0.8 & 5.3 & 6.2 & 5.7 & 6.6 & 5.7 & 5.8 & 4.7 & 6.4 \\ 
5-block insertion & 5.6 & 4.6 & 0.7 & 0.8 & 5.0 & 5.9 & 5.2 & 6.0 & 5.4 & 5.6 & 4.5 & 6.1 \\ 
6-block insertion & 5.3 & 4.5 & 0.7 & 0.8 & 4.7 & 5.6 & 4.9 & 5.6 & 5.1 & 5.3 & 4.3 & 5.8 \\ 
7-block insertion & 5.2 & 4.3 & 0.7 & 0.8 & 4.6 & 5.4 & 4.6 & 5.3 & 4.9 & 5.1 & 4.1 & 5.6 \\ 
8-block insertion & 5.0 & 4.3 & 0.7 & 0.8 & 4.4 & 5.2 & 4.4 & 5.0 & 4.7 & 4.9 & 3.9 & 5.3 \\ 
9-block insertion & 4.9 & 4.2 & 0.7 & 0.7 & 4.2 & 5.0 & 4.1 & 4.8 & 4.5 & 4.6 & 3.8 & 5.1 \\ 
10-block insertion & 4.7 & 4.0 & 0.6 & 0.7 & 4.0 & 4.8 & 3.9 & 4.6 & 4.3 & 4.4 & 3.6 & 4.9 \\ 
11-block insertion & 4.6 & 3.8 & 0.6 & 0.7 & 3.9 & 4.6 & 3.8 & 4.4 & 4.1 & 4.2 & 3.5 & 4.7 \\ 
12-block insertion & 4.4 & 3.7 & 0.6 & 0.7 & 3.7 & 4.4 & 3.6 & 4.2 & 3.9 & 4.0 & 3.3 & 4.5 \\ 
13-block insertion & 4.2 & 3.5 & 0.6 & 0.7 & 3.6 & 4.2 & 3.5 & 4.1 & 3.8 & 3.9 & 3.1 & 4.3 \\ 
Swap & 3.1 & 2.5 & 0.4 & 0.4 & 3.1 & 3.5 & 3.9 & 4.2 & 3.0 & 3.0 & 2.5 & 3.4 \\ \hline
\multicolumn{1}{c}{Mean} & 5.1 & 4.3 & 0.7 & 0.8 & 4.6 & 5.4 & 4.9 & 5.7 & 4.9 & 5.0 & 4.1 & 5.5 \\ \hline
\end{tabular}
\label{Time}
\end{table}

\begin{table}[!htbp]
\caption{Average time in seconds spent by each neighborhood in ILS-RVND$_{{\it Fast}}$}
\centering
\scriptsize
\onehalfspacing
\begin{tabular}{lcccccccccccc}
\hline
\multicolumn{1}{l}{} & \multicolumn{12}{c}{Group} \\ \cline{2-13}
\multicolumn{1}{c}{\up{Neighborhoods}} & 1 & 2 & 3 & 4 & 5 & 6 & 7 & 8 & 9 & 10 & 11 & 12 \\ \hline
1-block insertion & 1.2 & 0.6 & 0.2 & 0.1 & 1.8 & 1.1 & 2.3 & 1.6 & 1.2 & 0.9 & 1.1 & 0.9 \\ 
2-block insertion & 0.9 & 0.6 & 0.2 & 0.1 & 1.7 & 1.0 & 2.3 & 1.5 & 1.4 & 0.9 & 1.3 & 1.0 \\ 
3-block insertion & 0.7 & 0.5 & 0.1 & 0.1 & 1.4 & 0.9 & 2.2 & 1.3 & 1.6 & 1.0 & 1.4 & 1.0 \\ 
4-block insertion & 0.7 & 0.5 & 0.2 & 0.1 & 1.3 & 0.9 & 2.1 & 1.2 & 1.6 & 1.0 & 1.5 & 1.0 \\ 
5-block insertion & 0.7 & 0.4 & 0.2 & 0.1 & 1.2 & 0.8 & 1.8 & 1.1 & 1.5 & 1.0 & 1.4 & 1.0 \\ 
6-block insertion & 0.6 & 0.4 & 0.2 & 0.1 & 1.1 & 0.8 & 1.8 & 1.0 & 1.4 & 1.0 & 1.3 & 0.9 \\ 
7-block insertion & 0.6 & 0.4 & 0.2 & 0.1 & 1.1 & 0.8 & 1.8 & 1.0 & 1.3 & 0.9 & 1.3 & 0.9 \\ 
8-block insertion & 0.6 & 0.4 & 0.2 & 0.1 & 1.1 & 0.7 & 1.8 & 1.0 & 1.4 & 0.9 & 1.3 & 0.9 \\ 
9-block insertion & 0.6 & 0.4 & 0.2 & 0.1 & 1.2 & 0.7 & 1.9 & 1.1 & 1.4 & 0.9 & 1.3 & 0.9 \\ 
10-block insertion & 0.6 & 0.4 & 0.2 & 0.1 & 1.2 & 0.7 & 1.8 & 1.1 & 1.5 & 0.9 & 1.4 & 0.9 \\ 
11-block insertion & 0.5 & 0.3 & 0.2 & 0.1 & 1.3 & 0.8 & 1.9 & 1.1 & 1.5 & 0.9 & 1.5 & 0.9 \\ 
12-block insertion & 0.5 & 0.3 & 0.2 & 0.1 & 1.3 & 0.8 & 1.8 & 1.1 & 1.6 & 0.9 & 1.5 & 1.0 \\ 
13-block insertion & 0.6 & 0.3 & 0.2 & 0.1 & 1.4 & 0.8 & 1.9 & 1.2 & 1.7 & 1.0 & 1.5 & 1.0 \\ 
Swap & 0.7 & 0.3 & 0.1 & 0.1 & 0.9 & 0.5 & 1.2 & 0.7 & 0.6 & 0.5 & 0.5 & 0.5 \\ \hline
\multicolumn{1}{c}{Mean} & 0.7 & 0.4 & 0.2 & 0.1 & 1.3 & 0.8 & 1.9 & 1.1 & 1.4 & 0.9 & 1.3 & 0.9 \\ \hline
\end{tabular}
\label{TimeFast}
\end{table}

Finally, Table \ref{ILS-RVNDvsILS-RVND_Fast} compares the overall results found by ILS-RVND against those of ILS-RVND$_{Fast}$. 
We did not report the gap values for Groups 3 and 4 because the optimal solution for all instances, except for instance 24, is zero. The geometric means of Group 7 were also not reported in the table because the average gap of almost all instances were zero. In this last analysis we can observe that the speedup achieved by ILS-RVND$_{Fast}$ does not come at the expense of solution quality. The results demonstrate that there is no clear difference between both algorithms when it comes to the average gap between the average/best solutions and the optimal solution.

\begin{table}[!htbp]
\caption{ILS-RVND results vs ILS-RVND$_{Fast}$ results}
\centering
\scriptsize
\onehalfspacing
\setlength{\tabcolsep}{1.6mm}
\begin{tabular}{clccccccccccccc}
\hline
\multicolumn{1}{l}{} &  & \multicolumn{11}{c}{Group}  \\ \cline{3-14}
\multicolumn{1}{l}{} &  & 1 & 2 & 3 & 4 & 5 & 6 & 7 & 8 & 9 & 10 & 11 & 12 & \multicolumn{1}{c}{\up{Mean}} \\ 
\hline\noalign{\smallskip}
\multicolumn{1}{c|}{Arithm. Mean of} & ILS-RVND & 0.02 & 0.26 &  -- & -- & 0.00 & 0.00 & 0.00 & 0.00 & 0.00 & 0.00 & 0.00 & 0.00 & 0.21 \\ 
\multicolumn{1}{c|}{Best Gap ($\%$)} & ILS-RVND$_{Fast}$ & 0.07 & 0.45 &  -- & -- & 0.00 & 0.00 & 0.00 & 0.00 & 0.00 & 0.00 & 0.00 & 0.00 & 0.06 \\ 
 &  &  &  &  &  &  &  &  &  &  &  &  \\ 
\multicolumn{1}{c|}{Geom. Mean of} & ILS-RVND & 0.32 & 0.90 &  -- & -- & 0.03 & 0.24 & -- & 0.19 & 0.00 & 0.03 & 0.00 & 0.02 & 0.12 \\ 
\multicolumn{1}{c|}{Avg. Gap ($\%$)} & ILS-RVND$_{Fast}$ & 0.30 & 1.02 & -- & -- & 0.04 & 0.10 & -- & 0.18 & 0.00 & 0.02 & 0.00 & 0.01 & 0.08 \\ 
 &  &  &  &  &  &  &  &  &  &  &  &  \\ 
\multicolumn{1}{c|}{Geom. Mean of} & ILS-RVND & 0.84 & 2.00 &  -- & -- & 0.13 & 0.97 & -- & 1.17 & 0.01 & 0.08 & 0.02 & 0.15 & 0.40 \\ 
\multicolumn{1}{c|}{Worst Gap ($\%$)} & ILS-RVND$_{Fast}$ & 0.72 & 2.07 & -- & -- & 0.33 & 0.49 & -- & 0.92 & 0.01 & 0.10 & 0.01 & 0.07 & 0.31 \\ 
 &  &  &  &  &  &  &  &  &  &  &  &  \\ 
\multicolumn{1}{c|}{Arithm. Mean of} & ILS-RVND & 71.2 & 59.8 & 9.6 & 10.6 & 64.7 & 76.2 & 68.7 & 79.6 & 68.1 & 70.2 & 57.1 & 77.4 & 69.3 \\ 
\multicolumn{1}{c|}{Avg. Times (s)} & ILS-RVND$_{Fast}$ & 10.0 & 5.9 & 2.5 & 1.7 & 18.5 & 11.5 & 26.2 & 16.3 & 19.7 & 12.5 & 18.4 & 13.8 & 13.1 \\ \hline
\end{tabular}
\label{ILS-RVNDvsILS-RVND_Fast}
\end{table}

\subsection{Impact of the proposed limitation strategy on the performance of other metaheuristics}
\label{sec:metaheuristics}

In order to validate the generality of the proposed limitation strategy, we performed experiments with other local search based metaheuristics, namely GRASP and VNS. In the section we present the results obtained by such metaheuristics with and without the addition of such strategy. 

\subsubsection{GRASP}
\label{sec:metaheuristicsGRASP}

GRASP \citep{FeoResende1995} is a multi-start metaheuristic where at each restart a solution is generated using a greedy randomized approach and then possibly improved by a local search procedure. The level of greediness/randomness of the constructive phase is controlled by a parameter $0 \leq \alpha \leq 1$. In our implementation the value of $\alpha$ was chosen at random from the set $\{0.0, 0.1, 0.2, 0.3, 0.4, 0.5\}$. The local search is performed by the RVND procedure with the same neighborhoods used in the ILS algorithm. The number of restarts was set to 5000, where the first 250 restarts are associated with the learning phase, which is equivalent to 5\% of the number of restarts, as in ILS, while the value of $\theta$ was set to the same used in ILS, i.e., 0.90. 

Table \ref{tab:GRASP} shows the average results of 10 runs found by GRASP without and with the limitation strategy (GRASP$_{Fast}$). We can observe that GRASP$_{Fast}$ was, on average, roughly 2 times faster than GRASP without significantly affecting the quality of the solutions obtained, thus illustrating the effectiveness of the limitation strategy when considering a standard and non-tailored implementation of the metaheuristic.

\begin{table}[!htbp]
\caption{GRASP results vs GRASP$_{Fast}$ results}
\centering
\scriptsize
\onehalfspacing
\setlength{\tabcolsep}{0.5mm}
\begin{tabular}{clccccccccccccc}
\hline
\multicolumn{1}{l}{} &  & \multicolumn{11}{c}{Group}  \\ \cline{3-14}
\multicolumn{1}{l}{} &  & 1 & 2 & 3 & 4 & 5 & 6 & 7 & 8 & 9 & 10 & 11 & 12 & \multicolumn{1}{c}{\up{Mean}} \\ 
\hline\noalign{\smallskip}
\multicolumn{1}{c|}{Arithm. Mean of} & GRASP & 3.07 & 12.70 & -- & -- & 0.26 & 2.68 & 0.11 & 2.64 & 0.04 & 0.36 & 0.02 & 0.37 & 2.22 \\ 
\multicolumn{1}{c|}{Best Gap ($\%$)} & GRASP$_{Fast}$ & 3.07 & 11.61 & -- & -- & 0.28 & 1.80 & 0.04 & 3.01 & 0.03 & 0.38 & 0.02 & 0.27 & 2.05 \\ 
 &  &  &  &  &  &  &  &  &  &  &  &  &  &  \\ 
\multicolumn{1}{c|}{Geom. Mean of} & GRASP & 5.21 & 15.45 & -- & -- & 0.81 & 3.99 & 0.23 & 4.61 & 0.13 & 0.60 & 0.08 & 0.62 & 3.17 \\ 
\multicolumn{1}{c|}{Avg. Gap ($\%$)} & GRASP$_{Fast}$ & 4.92 & 16.81 & -- & -- & 0.87 & 3.92 & 0.28 & 4.61 & 0.12 & 0.57 & 0.08 & 0.57 & 3.27 \\ 
 &  &  &  &  &  &  &  &  &  &  &  &  &  &  \\ 
\multicolumn{1}{c|}{Geom. Mean of} & GRASP & 7.39 & 21.17 & -- & -- & 1.32 & 5.41 & 0.53 & 6.80 & 0.25 & 0.92 & 0.17 & 0.94 & 4.49 \\ 
\multicolumn{1}{c|}{Worst Gap ($\%$)} & GRASP$_{Fast}$ & 7.19 & 22.03 & -- & -- & 1.39 & 5.67 & 0.73 & 6.93 & 0.22 & 0.92 & 0.16 & 0.88 & 4.61 \\ 
 &  &  &  &  &  &  &  &  &  &  &  &  &  &  \\ 
\multicolumn{1}{c|}{Arithm. Mean of} & GRASP & 93.2 & 73.6 & 20.9 & 13.7 & 134.1 & 126.8 & 185.7 & 164.2 & 153.5 & 115.3 & 138.5 & 136.2 & 113.0 \\ 
\multicolumn{1}{c|}{Avg. Times (s)} & GRASP$_{Fast}$ & 50.1 & 24.6 & 15.5 & 6.8 & 91.3 & 58.4 & 138.9 & 84.1 & 105.5 & 51.4 & 95.0 & 64.0 & 65.5 \\ 
 &  &  &  &  &  &  &  &  &  &  &  &  &  &  \\ 
\multicolumn{1}{c|}{Avg. of time spend by}  & GRASP & 6.3 & 5.0 & 1.4 & 0.9 & 9.2 & 8.7 & 12.9 & 11.4 & 10.6 & 7.9 & 9.6 & 9.4 & 7.8 \\ 
\multicolumn{1}{c|}{the neighborhood (s)} & GRASP$_{Fast}$ & 3.2 & 1.5 & 1.0 & 0.4 & 6.2 & 3.8 & 9.5 & 5.6 & 7.2 & 3.4 & 6.5 & 4.2 & 4.4 \\ 
 &  &  &  &  &  &  &  &  &  &  &  &  &  &  \\ 
\multicolumn{ 2}{l}{Avg. of moves not evaluated ($\%$)} & 66.2 & 86.6 & 48.6 & 68.1 & 46.5 & 71.5 & 37.2 & 64.8 & 43.8 & 71.8 & 43.7 & 69.1 & 59.8 \\ 
 &  &  &  &  &  &  &  &  &  &  &  &  &  &  \\ 
\multicolumn{ 2}{l}{Avg. of improving moves not evaluated ($\%$)} & 10.6 & 9.2 & 3.5 & 1.9 & 13.1 & 13.4 & 15.4 & 15.5 & 14.2 & 14.5 & 14.5 & 14.5 & 11.7 \\ \hline
\end{tabular}
\label{tab:GRASP}
\end{table}

\subsubsection{VNS}
\label{sec:metaheuristicsVNS}

VNS \citep{MladenovicHansen1997} is a local search based metaheuristic that alternates between local search (intensification) and perturbation (diversification) procedures. The local optimal solutions are perturbed by means of one of the existing neighborhood operators. A local search is then applied over this perturbed solution. If the local optimal solution is not improved, a different neighborhood is used to perturb such incumbent solution. RVND was used as the local search procedure and the total number of iterations was set to 1000, where the first 5\% (50 iterations) are related to the learning phase. The value of $\theta$ and the neighborhoods were the same adopted in ILS and GRASP.

The average results of 10 runs obtained by VNS without and with the limitation strategy (VNS$_{Fast}$) can be found in Table \ref{tab:VNS}. There were practically no difference in terms of solution quality between VNS and VNS$_{Fast}$, but the latter was approximately 2 times faster than the first. The results suggest that the proposed limitation strategy was also effective for a straightforward implementation of this metaheuristic.

\begin{table}[!htbp]
\caption{VNS results vs VNS$_{Fast}$ results}
\centering
\scriptsize
\onehalfspacing
\setlength{\tabcolsep}{0.8mm}
\begin{tabular}{clccccccccccccc}
\hline
\multicolumn{1}{l}{} &  & \multicolumn{11}{c}{Group}  \\ \cline{3-14}
\multicolumn{1}{l}{} &  & 1 & 2 & 3 & 4 & 5 & 6 & 7 & 8 & 9 & 10 & 11 & 12 & \multicolumn{1}{c}{\up{Mean}} \\ 
\hline\noalign{\smallskip}
\multicolumn{1}{c|}{Arithm. Mean of} & VNS & 0.05 & 0.94 & -- & -- & 0.00 & 0.00 & 0.00 & 0.00 & 0.00 & 0.01 & 0.00 & 0.00 & 0.10 \\ 
\multicolumn{1}{c|}{Best Gap ($\%$)} & VNS$_{Fast}$ & 0.02 & 0.29 & -- & -- & 0.00 & 0.00 & 0.00 & 0.04 & 0.00 & 0.00 & 0.00 & 0.00 & 0.04 \\ 
 &  &  &  &  &  &  &  &  &  &  &  &  &  &  \\ 
\multicolumn{1}{c|}{Geom. Mean of} & VNS & 0.85 & 1.74 & -- & -- & 0.13 & 0.67 & 0.08 & 0.96 & 0.02 & 0.12 & 0.02 & 0.14 & 0.47 \\ 
\multicolumn{1}{c|}{Avg. Gap ($\%$)} & VNS$_{Fast}$ & 0.95 & 2.07 & -- & -- & 0.22 & 0.73 & 0.14 & 1.37 & 0.02 & 0.10 & 0.02 & 0.11 & 0.57 \\ 
 &  &  &  &  &  &  &  &  &  &  &  &  &  &  \\ 
\multicolumn{1}{c|}{Geom. Mean of} & VNS & 2.60 & 3.95 & -- & -- & 0.39 & 1.78 & 0.40 & 2.96 & 0.06 & 0.38 & 0.06 & 0.49 & 1.31 \\ 
\multicolumn{1}{c|}{Worst Gap ($\%$)} & VNS$_{Fast}$ & 2.93 & 6.94 & -- & -- & 0.66 & 1.73 & 0.55 & 4.04 & 0.08 & 0.33 & 0.06 & 0.39 & 1.77 \\ 
 &  &  &  &  &  &  &  &  &  &  &  &  &  &  \\ 
\multicolumn{1}{c|}{Arithm. Mean of} & VNS & 80.4 & 61.4 & 13.3 & 10.9 & 93.6 & 94.2 & 100.1 & 96.1 & 102.5 & 88.7 & 92.7 & 98.7 & 77.7 \\ 
\multicolumn{1}{c|}{Avg. Times (s)} & VNS$_{Fast}$ & 34.1 & 13.7 & 7.4 & 4.0 & 56.4 & 38.2 & 64.3 & 42.8 & 57.9 & 32.9 & 53.2 & 38.2 & 36.9 \\ 
 &  &  &  &  &  &  &  &  &  &  &  &  &  &  \\ 
\multicolumn{1}{c|}{Avg. of time spend by} & VNS & 5.7 & 4.4 & 0.9 & 0.8 & 6.7 & 6.7 & 7.1 & 6.8 & 7.3 & 6.3 & 6.6 & 7.0 & 5.5 \\ 
\multicolumn{1}{c|}{the neighborhood (s)} & VNS$_{Fast}$ & 2.4 & 1.0 & 0.5 & 0.3 & 4.0 & 2.7 & 4.6 & 3.0 & 4.1 & 2.3 & 3.8 & 2.7 & 2.6 \\ 
 &  &  &  &  &  &  &  &  &  &  &  &  &  &  \\ 
\multicolumn{ 2}{l}{Avg. of moves not evaluated ($\%$)} & 74.4 & 95.8 & 70.2 & 84.0 & 51.0 & 73.2 & 44.9 & 69.0 & 53.9 & 76.0 & 52.4 & 74.5 & 68.3 \\ 
 &  &  &  &  &  &  &  &  &  &  &  &  &  &  \\ 
\multicolumn{ 2}{l}{Avg. of improving moves not evaluated ($\%$)} & 10.3 & 9.7 & 11.4 & 12.5 & 11.5 & 12.8 & 12.1 & 13.8 & 12.1 & 14.0 & 12.2 & 13.8 & 12.2 \\ \hline
\end{tabular}
\label{tab:VNS}
\end{table}

\subsection{Results for problem $1|s_{ij}|\sum T_j$}
\label{sec:withoutWeight}

With a view of better assessing the robustness of the limitation strategy, we decided to test ILS-RVND and ILS-RVND$_{Fast}$ on benchmark instances of problem $1|s_{ij}|\sum T_j$, namely those generated by \citet{Rubin1995}, containing 32 instances ranging from 15 to 45 jobs, and those suggested by \citet{Gagne2002}, also containing 32 instances but ranging from 55 to 85 jobs. In this case the configuration adopted after tuning the parameters by using the same rationale presented in Section \ref{sec:tuning} was $L  = \{1,\dots,13\}$ + swap and $\theta = 0.75$.

We compare the results found by ILS-RVND$_{Fast}$ not only with \textbf{GVNS} \citep{Kirlik2012}, \textbf{ILS-RVND$_{\text{SBP}}$} \citep{Subramanian2014} and \textbf{LOX$\oplus$B} \citep{Xu2014}, where in the latter the authors only reported the best of 100 executions, but also with the algorithms listed below along with the type of result presented.

\textbf{ACO$_{\text{GPG}}$}: Ant Colony of \citet{Gagne2002}. Best, worse, average and median of 20 runs for the instances proposed by the authors. Furthermore, the best solutions found by ACO$_{\text{GPG}}$ for the instances of \citet{Rubin1995} were reported by \citet{Liao2007}.

\textbf{Tabu$_{\text{GGP}}$}: TS and VNS  of \citet{Gagne2005}. Best of 10 runs for the instances of \citet{Gagne2002}, while the results found for the instances of  \citet{Rubin1995} were reported by \citet{Ying2009}.

\textbf{GRASP$_{\text{GS}}$}: GRASP of \citet{Gupta2006}. Best, worse, average and median of 20 runs for the instances of \citet{Gagne2002}.

\textbf{ACO$_{\text{LJ}}$}: Ant Colony of \citet{Liao2007}. Best of 10 runs for all instances.

\textbf{ILS$_{\text{ANK}}$}: Iterated Local Search of \citet{Arroyo2009}. Best, worse and average of 20 runs for the instances of \citet{Gagne2002}.

\textbf{IG}: Iterated Greedy of \citet{Ying2009}. Best of 10 runs for all instances.
 
\textbf{Opt}: Optimal solution found by the exact algorithm of \citet{Tanaka2013}, except for instances 851 and 855. For these instances, the authors provide a lower bound of 357 and 254, respectively.\\

Tables \ref{tab:Comparison2} and \ref{tab:Comparison3} compare the results found by ILS-RVND$_{Fast}$ with the best known methods found in the literature. The proposed algorithm was capable of finding the optimal solutions of all cases, except for instances 751, 851 and 855. The average computational time was 0.7 seconds for the instances of \citet{Rubin1995} and 12.5 seconds for the instances of \citet{Gagne2002}.

\begin{table}[!htbp]
\caption{Results for the instances of \citet{Rubin1995}}
\centering
\onehalfspacing
\scriptsize
\setlength{\tabcolsep}{0.41mm}
\setlength{\LTcapwidth}{10in}
\begin{tabular}{cccccccccccccccccccccr}
\hline
 &  &  & ACO$_{\text{GPG}}$ & & Tabu$_{\text{GGP}}$ & & ACO$_{\text{LJ}}$ & & IG & & GVNS & & \multicolumn{2}{c}{ILS-RVND$_{\text{SBP}}$} & & LOX$\oplus$B & & \multicolumn{4}{c}{ILS-RVND$_{Fast}$} \\ \cline{4-4}\cline{6-6}\cline{8-8}\cline{10-10}\cline{12-12}\cline{14-15}\cline{17-17}\cline{19-22}
 Inst. & $n$ & Opt. &  &  &  &  &  &  &  &  &  &  &  &  &  &  &  &  &  &  & \multicolumn{1}{c}{Avg. Time} \\ 
 &  &  & \up{Best} &  & \up{Best} &  & \up{Best} &  & \up{Best} &  & \up{Best} &  & \up{Best} & \up{Avg.} &  & \up{Best} &  & \up{Best} & \up{Avg.} & \up{Worst} & \multicolumn{1}{c}{(s)} \\  \hline  \noalign{\smallskip}
401 & 15 & 90 & 90 &  & 90 &  & 90 &  & 90 &  & 90 &  & 90 & 90.0 &  & 90 &  & 90 & 90 & 90 & $<$ 0.1 \\ 
402 & 15 & 0 & 0 &  & 0 &  & 0 &  & 0 &  & 0 &  & 0 & 0.0 &  & 0 &  & 0 & 0 & 0 & $<$ 0.1 \\ 
403 & 15 & 3418 & 3418 &  & 3418 &  & 3418 &  & 3418 &  & 3418 &  & 3418 & 3418.0 &  & 3418 &  & 3418 & 3418 & 3418 & 0.1 \\ 
404 & 15 & 1067 & 1067 &  & 1067 &  & 1067 &  & 1067 &  & 1067 &  & 1067 & 1067.0 &  & 1067 &  & 1067 & 1067 & 1067 & 0.1 \\ 
405 & 15 & 0 & 0 &  & 0 &  & 0 &  & 0 &  & 0 &  & 0 & 0.0 &  & 0 &  & 0 & 0 & 0 & $<$ 0.1 \\
406 & 15 & 0 & 0 &  & 0 &  & 0 &  & 0 &  & 0 &  & 0 & 0.0 &  & 0 &  & 0 & 0 & 0 & $<$ 0.1 \\ 
407 & 15 & 1861 & 1861 &  & 1861 &  & 1861 &  & 1861 &  & 1861 &  & 1861 & 1861.0 &  & 1861 &  & 1861 & 1861 & 1861 & $<$ 0.1 \\ 
408 & 15 & 5660 & 5660 &  & 5660 &  & 5660 &  & 5660 &  & 5660 &  & 5660 & 5660.0 &  & 5660 &  & 5660 & 5660 & 5660 & 0.1 \\ \hline  \noalign{\smallskip}
501 & 25 & 261 & 261 &  & 261 &  & 263 &  & 261 &  & 261 &  & 261 & 261.0 &  & 261 &  & 261 & 261 & 261 & 0.2 \\ 
502 & 25 & 0 & 0 &  & 0 &  & 0 &  & 0 &  & 0 &  & 0 & 0.0 &  & 0 &  & 0 & 0 & 0 & $<$ 0.1 \\
503 & 25 & 3497 & 3497 &  & 3503 &  & 3497 &  & 3497 &  & 3497 &  & 3497 & 3497.0 &  & 3497 &  & 3497 & 3497 & 3497 & 0.2 \\ 
504 & 25 & 0 & 0 &  & 0 &  & 0 &  & 0 &  & 0 &  & 0 & 0.0 &  & 0 &  & 0 & 0 & 0 & $<$ 0.1 \\ 
505 & 25 & 0 & 0 &  & 0 &  & 0 &  & 0 &  & 0 &  & 0 & 0.0 &  & 0 &  & 0 & 0 & 0 & $<$ 0.1 \\
506 & 25 & 0 & 0 &  & 0 &  & 0 &  & 0 &  & 0 &  & 0 & 0.0 &  & 0 &  & 0 & 0 & 0 & $<$ 0.1 \\ 
507 & 25 & 7225 & 7268 &  & 7225 &  & 7225 &  & 7225 &  & 7225 &  & 7225 & 7225.0 &  & 7225 &  & 7225 & 7225 & 7225 & 0.2 \\ 
508 & 25 & 1915 & 1945 &  & 1915 &  & 1915 &  & 1915 &  & 1915 &  & 1915 & 1915.0 &  & 1915 &  & 1915 & 1915 & 1915 & 0.4 \\  \hline  \noalign{\smallskip}
601 & 35 & 12 & 16 &  & 12 &  & 14 &  & 12 &  & 12 &  & 12 & 12.7 &  & 12 &  & 12 & 12 & 12 & 0.6 \\ 
602 & 35 & 0 & 0 &  & 0 &  & 0 &  & 0 &  & 0 &  & 0 & 0.0 &  & 0 &  & 0 & 0 & 0 & $<$ 0.1 \\ 
603 & 35 & 17587 & 17685 &  & 17605 &  & 17654 &  & 17587 &  & 17587 &  & 17587 & 17590.6 &  & 17587 &  & 17587 & 17589.0 & 17607 & 0.7 \\ 
604 & 35 & 19092 & 19213 &  & 19168 &  & 19092 &  & 19092 &  & 19092 &  & 19092 & 19092.9 &  & 19092 &  & 19092 & 19092.9 & 19101 & 1.3 \\ 
605 & 35 & 228 & 247 &  & 228 &  & 240 &  & 228 &  & 228 &  & 228 & 229.2 &  & 228 &  & 228 & 228.6 & 229 & 0.7 \\ 
606 & 35 & 0 & 0 &  & 0 &  & 0 &  & 0 &  & 0 &  & 0 & 0.0 &  & 0 &  & 0 & 0 & 0 & $<$ 0.1 \\ 
607 & 35 & 12969 & 13088 &  & 12969 &  & 13010 &  & 12969 &  & 12969 &  & 12969 & 12974.2 &  & 12969 &  & 12969 & 12969 & 12969 & 0.7 \\ 
608 & 35 & 4732 & 4733 &  & 4732 &  & 4732 &  & 4732 &  & 4732 &  & 4732 & 4732.0 &  & 4732 &  & 4732 & 4732 & 4732 & 1.6 \\ \hline  \noalign{\smallskip}
701 & 45 & 97 & 103 &  & 98 &  & 103 &  & 103 &  & 99 &  & 97 & 99.6 &  & 97 &  & 97 & 98.1 & 100 & 1.4 \\ 
702 & 45 & 0 & 0 &  & 0 &  & 0 &  & 0 &  & 0 &  & 0 & 0.0 &  & 0 &  & 0 & 0 & 0 & $<$ 0.1 \\ 
703 & 45 & 26506 & 26663 &  & 26506 &  & 26568 &  & 26496$^1$ &  & 26506 &  & 26506 & 26508.8 &  & 26506 &  & 26506 & 26508.1 & 26519 & 1.9 \\ 
704 & 45 & 15206 & 15495 &  & 15213 &  & 15409 &  & 15206 &  & 15206 &  & 15206 & 15206.0 &  & 15206 &  & 15206 & 15206 & 15206 & 3.9 \\ 
705 & 45 & 200 & 222 &  & 200 &  & 219 &  & 200 &  & 202 &  & 200 & 203.6 &  & 200 &  & 200 & 202.4 & 204 & 1.6 \\ 
706 & 45 & 0 & 0 &  & 0 &  & 0 &  & 0 &  & 0 &  & 0 & 0.0 &  & 0 &  & 0 & 0 & 0 & $<$ 0.1 \\ 
707 & 45 & 23789 & 24017 &  & 23804 &  & 23931 &  & 23794 &  & 23789 &  & 23789 & 23790.5 &  & 23789 &  & 23789 & 23789 & 23789 & 2.3 \\ 
708 & 45 & 22807 & 23351 &  & 22873 &  & 23028 &  & 22807 &  & 22807 &  & 22807 & 22807.0 &  & 22807 &  & 22807 & 22807 & 22807 & 3.7 \\ \hline
 &  &  &  &  &  &  &  &  &  &  &  &  &  &  &  &  &  &  &  & Avg. & 0.7 \\ \hline
\multicolumn{21}{l}{\tiny $^1$: Value smaller than the optimum reported by \citet{Tanaka2013}.} \\
\end{tabular}
\label{tab:Comparison2}
\end{table}

\begin{table}[!htbp]
\caption{Results for the instances of \citet{Gagne2002}}
\centering
\onehalfspacing
\scriptsize
\setlength{\tabcolsep}{0.51mm}
\setlength{\LTcapwidth}{10in}
\begin{tabular}{cccccccccccccccccccccr}
\hline
 &  &  & ACO$_{\text{GPG}}$ & & Tabu$_{\text{GGP}}$ & & ACO$_{\text{LJ}}$ & & IG & & GVNS & & \multicolumn{2}{c}{ILS-RVND$_{\text{SBP}}$} & & LOX$\oplus$B & & \multicolumn{4}{c}{ILS-RVND$_{Fast}$} \\ \cline{4-4}\cline{6-6}\cline{8-8}\cline{10-10}\cline{12-12}\cline{14-15}\cline{17-17}\cline{19-22}
Inst. & $n$ & Opt. &  &  &  &  &  &  &  &  &  &  &  &  &  &  &  &  &  &  & \multicolumn{1}{c}{Avg. Time} \\ 
 &  &  & \up{Best} &  & \up{Best} &  & \up{Best} &  & \up{Best} &  & \up{Best} &  & \up{Best} & \up{Avg.} &  & \up{Best} &  & \up{Best} & \up{Avg.} & \up{Worst} & \multicolumn{1}{c}{(s)} \\  \hline  \noalign{\smallskip}
551 & 55 & 183 & 212 &  & 185 &  & 183 &  & 183 &  & 194 &  & 185 & 193.1 &  & 183 &  & 183 & 189.8 & 194 & 2.8 \\ 
552 & 55 & 0 & 0 &  & 0 &  & 0 &  & 0 &  & 0 &  & 0 & 0.0 &  & 0 &  & 0 & 0 & 0 & $<$ 0.1 \\
553 & 55 & 40498 & 40828 &  & 40644 &  & 40676 &  & 40598 &  & 40540 &  & 40498 & 40533.5 &  & 40498 &  & 40498 & 40524.1 & 40583 & 3.8 \\ 
554 & 55 & 14653 & 15091 &  & 14711 &  & 14684 &  & 14653 &  & 14653 &  & 14653 & 14653.0 &  & 14653 &  & 14653 & 14653 & 14653 & 8.9 \\ 
555 & 55 & 0 & 0 &  & 0 &  & 0 &  & 0 &  & 0 &  & 0 & 0.0 &  & 0 &  & 0 & 0 & 0 & $<$ 0.1 \\ 
556 & 55 & 0 & 0 &  & 0 &  & 0 &  & 0 &  & 0 &  & 0 & 0.0 &  & 0 &  & 0 & 0 & 0 & $<$ 0.1 \\ 
557 & 55 & 35813 & 36489 &  & 35841 &  & 36420 &  & 35827 &  & 35830 &  & 35830 & 35837.5 &  & 35813 &  & 35813 & 35816.9 & 35834 & 5.0 \\ 
558 & 55 & 19871 & 20624 &  & 19872 &  & 19888 &  & 19871 &  & 19871 &  & 19871 & 19871.0 &  & 19871 &  & 19871 & 19871 & 19871 & 8.1 \\ \hline  \noalign{\smallskip}
651 & 65 & 247 & 295 &  & 268 &  & 268 &  & 268 &  & 264 &  & 259 & 268.2 &  & 247 &  & 247 & 258.7 & 262 & 5.6 \\ 
652 & 65 & 0 & 0 &  & 0 &  & 0 &  & 0 &  & 0 &  & 0 & 0.0 &  & 0 &  & 0 & 0 & 0 & $<$ 0.1 \\ 
653 & 65 & 57500 & 57779 &  & 57602 &  & 57584 &  & 57584 &  & 57515 &  & 57508 & 57556.1 &  & 57500 &  & 57500 & 57538.5 & 57603 & 7.8 \\
654 & 65 & 34301 & 34468 &  & 34466 &  & 34306 &  & 34306 &  & 34301 &  & 34301 & 34305.4 &  & 34301 &  & 34301 & 34302.6 & 34309 & 15.8 \\ 
655 & 65 & 0 & 13 &  & 2 &  & 7 &  & 2 &  & 4 &  & 4 & 6.0 &  & 2 &  & 0 & 2.3 & 4 & 5.2 \\ 
656 & 65 & 0 & 0 &  & 0 &  & 0 &  & 0 &  & 0 &  & 0 & 0.0 &  & 0 &  & 0 & 0 & 0 & $<$ 0.1 \\ 
657 & 65 & 54895 & 56246 &  & 55080 &  & 55389 &  & 55080 &  & 54895 &  & 54895 & 54942.8 &  & 54895 &  & 54895 & 54937.7 & 55042 & 8.9 \\ 
658 & 65 & 27114 & 29308 &  & 27187 &  & 27208 &  & 27114 &  & 27114 &  & 27114 & 27114.0 &  & 27114 &  & 27114 & 27114 & 27114 & 18.7 \\ \hline  \noalign{\smallskip}
751 & 75 & 225 & 263 &  & 241 &  & 241 &  & -- &  & 241 &  & 237 & 243.0 &  & 229 &  & 227 & 235.6 & 239 & 8.2 \\ 
752 & 75 & 0 & 0 &  & 0 &  & 0 &  & 0 &  & 0 &  & 0 & 0.0 &  & 0 &  & 0 & 0 & 0 & $<$ 0.1 \\ 
753 & 75 & 77544 & 78211 &  & 77739 &  & 77663 &  & 77663 &  & 77627 &  & 77559 & 77636.5 &  & 77544 &  & 77544 & 77575.3 & 77606 & 14.3 \\ 
754 & 75 & 35200 & 35826 &  & 35709 &  & 35630 &  & 35250 &  & 35219 &  & 35209 & 35227.7 &  & 35200 &  & 35200 & 35217.7 & 35239 & 32.4 \\ 
755 & 75 & 0 & 0 &  & 0 &  & 0 &  & 0 &  & 0 &  & 0 & 0.0 &  & 0 &  & 0 & 0 & 0 & $<$ 0.1 \\ 
756 & 75 & 0 & 0 &  & 0 &  & 0 &  & 0 &  & 0 &  & 0 & 0.0 &  & 0 &  & 0 & 0 & 0 & $<$ 0.1 \\ 
757 & 75 & 59635 & 61513 &  & 59763 &  & 60108 &  & 59763 &  & 59716 &  & 59644 & 59724.2 &  & 59635 &  & 59635 & 59696.2 & 59773 & 17.0 \\ 
758 & 75 & 38339 & 40277 &  & 38789 &  & 38704 &  & 38431 &  & 38339 &  & 38339 & 38369.5 &  & 38339 &  & 38339 & 38352.3 & 38426 & 34.2 \\ \hline  \noalign{\smallskip}
851 & 85 & 360$^1$ & 453 &  & 384 &  & 455 &  & 390 &  & 402 &  & 381 & 392.4 &  & 381 &  & 372 & 381.3 & 387 & 16.4 \\ 
852 & 85 & 0 & 0 &  & 0 &  & 0 &  & 0 &  & 0 &  & 0 & 0.0 &  & 0 &  & 0 & 0 & 0 & $<$ 0.1 \\ 
853 & 85 & 97497 & 98540 &  & 97880 &  & 98443 &  & 97880 &  & 97595 &  & 97497 & 97632.9 &  & 97497 &  & 97497 & 97534.1 & 97559 & 22.7 \\ 
854 & 85 & 79042 & 80693 &  & 80122 &  & 79553 &  & 79631 &  & 79271 &  & 79090 & 79187.9 &  & 79086 &  & 79042 & 79128.5 & 79218 & 58.5 \\ 
855 & 85 & 258$^1$ & 333 &  & 283 &  & 324 &  & 283 &  & 280 &  & 274 & 279.2 &  & 270 &  & 262 & 267.0 & 272 & 18.1 \\ 
856 & 85 & 0 & 0 &  & 0 &  & 0 &  & 0 &  & 0 &  & 0 & 0.0 &  & 0 &  & 0 & 0 & 0 & $<$ 0.1 \\
857 & 85 & 87011 & 89654 &  & 87244 &  & 87504 &  & 87244 &  & 87075 &  & 87064 & 87135.8 &  & 87011 &  & 87011 & 87049.1 & 87155 & 27.4 \\ 
858 & 85 & 74739 & 77919 &  & 75533 &  & 75506 &  & 75029 &  & 74755 &  & 74739 & 74783.4 &  & 74739 &  & 74739 & 74763.3 & 74792 & 62.8 \\ \hline
 &  &  &  &  &  &  &  &  &  &  &  &  &  &  &  &  &  &  &  & Avg. & 12.5 \\ \hline
 \multicolumn{21}{l}{\tiny $^1$: Best upper bound found by \citet{Tanaka2013}; optimality not proven.} \\
\end{tabular}
\label{tab:Comparison3}
\end{table}

A summary of the comparison between ILS-RVND$_{Fast}$ and the other methods from the literature is presented in Table \ref{tab:SummaryComparison2}. From this table, it can be observed that ILS-RVND$_{Fast}$ visibly outperforms the existing heuristics in  terms of solution quality. Moreover, our best and average solutions were never worse than those found by other heuristics.

\begin{table}[!htbp]
\caption{Summary of the results found by ILS-RVND$_{Fast}$ compared to several heuristic methods from the literature ($1|s_{ij}|\sum T_j$)}
\centering
\scriptsize
\onehalfspacing
\setlength{\tabcolsep}{0.8mm}
\begin{tabular}{lccccccccc}
\hline
 & ACO$_{\text{GPG}}$ & Tabu$_{\text{GGP}}$ & GRASP$_{\text{GS}}$ & ACO$_{\text{LJ}}$ & ILS$_{\text{ANK}}$ & IG & GVNS & ILS-RVND$_{\text{SBP}}$ & LOX$\oplus$B \\ \hline
$\#Best$ improved & 36 & 29 & 19 & 32 & 19 & 19 & 18 & 13 & 5 \\ 
$\#Best$ equaled & 28 & 35 & 13 & 32 & 13 & 43 & 46 & 51 & 59 \\ 
$\#Best$ worse & 0 & 0 & 0 & 0 & 0 & 0 & 0 & 0 & 0 \\ 
$\#Avg.$ better than the Best & 36 & 26 & 19 & 32 & 18 & 18 & 15 & 6 & 1 \\ 
$\#Avg.$ equal to the Best & 28 & 32 & 13 & 30 & 13 & 38 & 39 & 39 & 39 \\ 
$\#Avg.$ improved & 23 & -- & 23 & -- & 23 & -- & -- & 27 & -- \\ 
$\#Avg.$ equaled & 9 & -- & 9 & -- & 9 & -- & -- & 37 & -- \\ 
$\#Avg.$ worse & 0 & -- & 0 & -- & 0 & -- & -- & 0 & -- \\ 
$\#Worst$ better than the Best & 36 & 22 & 18 & 30 & 15 & 14 & 7 & 1 & 0 \\ 
$\#Worst$ equal to the Best & 28 & 32 & 13 & 30 & 13 & 38 & 41 & 40 & 39 \\ 
$\#Worst$ better than the Avg. & 23 & -- & 23 & -- & 23 & -- & -- & 12 & -- \\ 
$\#Worst$ equal to the Avg. & 9 & -- & 9 & -- & 9 & -- & -- & 36 & -- \\ 
$\#Worst$ improved & 23 & -- & 23 & -- & 23 & -- & -- & 11 & -- \\ 
$\#Worst$ equaled & 9 & -- & 9 & -- & 9 & -- & -- & 33 & -- \\ 
$\#Worst$ worse & 0 & -- & 0 & -- & 0 & -- & -- & 20 & -- \\ 
$\#$Reported values & 32 / 64$^1$ & 64 & 32 & 64 & 32 & 62$^2$ & 64 & 64 & 64 \\ \hline
\multicolumn{10}{l}{\tiny $^1$: 64 for the best solutions and 32 for the average and worse solutions.} \\
\multicolumn{10}{l}{\tiny $^2$: Results for the instance 751 not reported and the value reported for instance 703 is smaller than the optimum.} \\
\end{tabular}
\label{tab:SummaryComparison2}
\end{table}

Table \ref{ILS-RVNDvsILS-RVND_Fast2} summarizes the results found by ILS-RVND and ILS-RVND$_{Fast}$. The geometric mean of the instances involving 15 and 25 jobs were not reported because the gaps of both groups were zero. Regarding the quality of the solution obtained, both methods produced similar results, but ILS-RVND$_{Fast}$ was, on average, approximately 7 times faster than ILS-RVND.

\begin{table}[!htbp]
\caption{ILS-RVND results vs ILS-RVND$_{Fast}$ results ($1|s_{ij}|\sum T_j$)}
\centering
\scriptsize
\onehalfspacing
\setlength{\tabcolsep}{1.4mm}
\begin{tabular}{clccccccccc}
\hline
 &  & \multicolumn{7}{c}{$n$}  \\ \cline{3-10}
 &  & 15 & 25 & 35 & 45 & 55 & 65 & 75 & 85 & \multicolumn{1}{c}{Mean} \\ 
\hline\noalign{\smallskip}
\multicolumn{1}{c|}{Arithm. Mean of} & ILS-RVND & 0.00 & 0.00 & 0.00 & 0.00 & 0.34 & 0.30 & 0.34 & 1.01 & 0.25  \\ 
\multicolumn{1}{c|}{Best Gap ($\%$)} & ILS-RVND$_{Fast}$  & 0.00 & 0.00 & 0.00 & 0.00 & 0.00 & 0.00 & 0.11 & 0.61 & 0.09 \\ 
&  &  &  &  &  &  &  &  \\ 
\multicolumn{1}{c|}{Geom. Mean of} & ILS-RVND & -- & -- & 0.01 & 0.77 & 1.10 & 2.27 & 0.95 & 1.83 & 0.15  \\ 
\multicolumn{1}{c|}{Avg. Gap ($\%$)} & ILS-RVND$_{Fast}$  & -- & -- & 0.09 & 0.78 & 1.26 & 1.22 & 0.99 & 1.60 & 0.12  \\ 
&  &  &  &  &  &  &  &  \\ 
\multicolumn{1}{c|}{Geom. Mean of} & ILS-RVND  & -- & -- & 0.07 & 2.05 & 1.52 & 3.05 & 1.53 & 2.36 & 0.35  \\ 
\multicolumn{1}{c|}{Worst Gap ($\%$)} & ILS-RVND$_{Fast}$  & -- & -- & 0.20 & 1.71 & 2.09 & 1.64 & 1.37 & 2.24 & 0.34  \\ 
&  &  &  &  &  &  &  &  \\ 
\multicolumn{1}{c|}{Arithm. Mean of} & ILS-RVND  & 0.1 & 0.5 & 3.7 & 11.1 & 22.2 & 54.3 & 90.6 & 195.9 & 47.3   \\ 
\multicolumn{1}{c|}{Avg. Times (s)} & ILS-RVND$_{Fast}$  & $<$ 0.1 & 0.1 & 0.7 & 1.8 & 3.6 & 7.8 & 13.3 & 25.7 & 6.6 \\
&  &  &  &  &  &  &  &  \\ 
\multicolumn{1}{c|}{Avg. of time spend by} & ILS-RVND  & $<$ 0.1 & 0.1 & 0.6 & 1.8 & 3.7 & 9.0 & 15.1 & 32.6 & 7.8 \\ 
\multicolumn{1}{c|}{the neighborhood (s)} & ILS-RVND$_{Fast}$  & $<$ 0.1 & $<$ 0.1 & 0.1 & 0.3 & 0.7 & 1.5 & 2.5 & 4.7 & 1.2 \\
&  &  &  &  &  &  &  &  \\ 
\multicolumn{ 2}{l}{Avg. of moves not evaluated ($\%$)} & 83.8 & 91.4 & 94.2 & 94.6 & 93.6 & 93.8 & 93.3 & 95.0 & 92.5 \\
&  &  &  &  &  &  &  &  \\ 
\multicolumn{ 2}{l}{Avg. of improving moves not evaluated ($\%$)} & 22.1 & 13.5 & 18.3 & 18.5 & 16.2 & 18.6 & 16.7 & 18.9 & 17.8 \\ \hline
\end{tabular}
\label{ILS-RVNDvsILS-RVND_Fast2}
\end{table}

\section{Concluding Remarks}
\label{sec:conclusions}

In this paper we proposed a simple but very efficient local search limitation strategy for problem $1|s_{ij}|\sum w_jT_j$. This strategy aims at speeding up the local search process by avoiding evaluations of unpromising moves. The setup variation is used to estimate whether or not a move should be evaluated. In particular, a move is only evaluated if the setup variation is smaller than a given threshold value, which in turn depends on the characteristics of the instance and on the neighborhood structure. Therefore, instead of tuning a value for this threshold \textit{a priori}, we developed a procedure that automatically estimates a value for this parameter. Furthermore, we presented a detailed description of how the limitation strategy can be incorporated into the neighborhoods swap and $l$-block insertion. 

The proposed approach was embedded in the ILS-RVND algorithm \cite{Subramanian2012b}, which is a simple local search based metaheuristic that was successfully applied to many combinatorial optimization problems, including the $1|s_{ij}|\sum w_jT_j$ \cite{Subramanian2014}. This enhanced version of the algorithm was denoted as ILS-RVND$_{Fast}$. Extensive computational experiments were carried out on well-known benchmark instances and the results obtained suggest that ILS-RVND$_{Fast}$ is capable of producing extremely competitive results both in terms of average solutions and CPU time. When analyzing the impact of the limitation strategy, it was possible to confirm that the high speedups achieved did not come at the expense of solution quality. As a result, a considerable number of neighborhoods could be used without significantly increasing the CPU time, which was crucial for the high performance of ILS-RVND$_{Fast}$. 

We performed similar experiments with standard and non-tailored implementations of other well-known local search based metaheuristics, namely GRASP and VNS, and the versions with the limitation strategy were approximately 2 times faster, without significant loss in solution quality, than those without the inclusion of such strategy. Finally, we also performed experiments with ILS-RVND$_{Fast}$ on benchmark instances of problem $1|s_{ij}|\sum T_j$ and the results obtained ratified the effectiveness of the limitation strategy in a related problem. 

\section*{Acknowledgments}

The authors would like to thank Arthur Kramer for valuable discussions concerning the move evaluation schemes.
This research was partially supported by the Conselho Nacional de Desenvolvimento Cient{\'i}fico e Tecnol{\'o}gico ́(CNPq), grant 471158/2012-7.

 \bibliographystyle{mmsbib}
\bibliography{referencias}

\begin{thebibliography}{42}
\providecommand{\natexlab}[1]{#1}
\providecommand{\url}[1]{\texttt{#1}}
\expandafter\ifx\csname urlstyle\endcsname\relax
  \providecommand{\doi}[1]{doi: #1}\else
  \providecommand{\doi}{doi: \begingroup \urlstyle{rm}\Url}\fi

\bibitem[Anghinolfi and Paolucci(2008)]{Anghinolfi2008}
Anghinolfi, D. and Paolucci, M. (2008), A new ant colony optimization approach
  for the single machine total weighted tardiness scheduling problem.
\newblock \emph{International Journal of Operations Research}, v. 5, p.
  \penalty0 44--60.

\bibitem[Anghinolfi and Paolucci(2009)]{Anghinolfi2009}
Anghinolfi, D. and Paolucci, M. (2009), A new discrete particle swarm
  optimization approach for the single-machine total weighted tardiness
  scheduling problem with sequence-dependent setup times.
\newblock \emph{European Journal of Operational Research}, v. 193, \penalty0 n.
  1, p. \penalty0 73--85.

\bibitem[Arroyo \emph{et~al}.(2009)Arroyo, Nunes, and Kamke]{Arroyo2009}
Arroyo, J. E.~C., Nunes, G. V.~P. and Kamke, E.~H.
\newblock Iterative local search heuristic for the single machine scheduling
  problem with sequence dependent setup times and due dates.
\newblock Yu, G., K{\"o}ppen, M., Chen, S.-M. and Niu, X. (Eds.), \emph{HIS
  (1)}, p. 505--510. IEEE Computer Society, 2009.

\bibitem[Bo{\. z}ejko(2010)]{Bozejko2010}
Bo{\. z}ejko, W. (2010), Parallel path relinking method for the single machine
  total weighted tardiness problem with sequence-dependent setups.
\newblock \emph{Journal of Intelligent Manufacturing}, v. 21, \penalty0 n. 6,
  p. \penalty0 777--785.

\bibitem[Chao and Liao(2012)]{Chao2012}
Chao, C.-W. and Liao, C.-J. (2012), A discrete electromagnetism-like mechanism
  for single machine total weighted tardiness problem with sequence-dependent
  setup times.
\newblock \emph{Applied Soft Computing}, v. 12, \penalty0 n. 9, p. \penalty0
  3079--3087.

\bibitem[Cicirello(2006)]{Cicirello2006}
Cicirello, V.~A.
\newblock Non-wrapping order crossover: An order preserving crossover operator
  that respects absolute position.
\newblock \emph{Proceedings of the Genetic and Evolutionary Computation
  Conference}, p. 1125--1131. ACM Press, 2006.

\bibitem[Cicirello and Smith(2005)]{Cicirello2005}
Cicirello, V.~A. and Smith, S.~F. (2005), Enhancing stochastic search
  performance by value-biased randomization of heuristics.
\newblock \emph{Journal of Heuristics}, v. 11, p. \penalty0 5--34.

\bibitem[Deng and Gu(2014)]{Deng2014}
Deng, G. and Gu, X. (2014), An iterated greedy algorithm for the single-machine
  total weighted tardiness problem with sequence-dependent setup times.
\newblock \emph{International Journal of Systems Science}, v. 45, \penalty0 n.
  3, p. \penalty0 351--362.

\bibitem[Feo and Resende(1995)]{FeoResende1995}
Feo, T.~A. and Resende, M. G.~C. (1995), Greedy randomized adaptive search
  procedures.
\newblock \emph{Journal of Global Optimization}, v. 6, \penalty0 n. 2, p.
  \penalty0 109--133.

\bibitem[Gagn{\'e} \emph{et~al}.(2005)Gagn{\'e}, Gravel, and Price]{Gagne2005}
Gagn{\'e}, C., Gravel, M. and Price, W.~L. (2005), Using metaheuristic
  compromise programming for the solution of multiple-objective scheduling
  problems.
\newblock \emph{Journal of the Operational Research Society}, v. 56, \penalty0
  n. 6, p. \penalty0 687--698.

\bibitem[Gagn{\'e} \emph{et~al}.(2002)Gagn{\'e}, Price, and Gravel]{Gagne2002}
Gagn{\'e}, C., Price, W.~L. and Gravel, M. (2002), Comparing an aco algorithm
  with other heuristics for the single machine scheduling problem with
  sequence-dependent setup times.
\newblock \emph{The Journal of the Operational Research Society}, v. 53,
  \penalty0 n. 8, p. \penalty0 895--906.

\bibitem[Graham \emph{et~al}.(1979)Graham, Lawler, Lenstra, and
  Kan]{Graham1979}
Graham, R., Lawler, E., Lenstra, J. and Kan, A.
\newblock Optimization and approximation in deterministic sequencing and
  scheduling: a survey.
\newblock P.L.~Hammer, E.~J. and Korte, B. (Eds.), \emph{Discrete Optimization
  II Proceedings of the Advanced Research Institute on Discrete Optimization
  and Systems Applications of the Systems Science Panel of NATO and of the
  Discrete Optimization Symposium co-sponsored by IBM Canada and SIAM Banff,
  Aha. and Vancouver}, volume~5 of \emph{Annals of Discrete Mathematics}, p.
  287--326. Elsevier, 1979.

\bibitem[Guo and Tang(2015)]{Guo2015}
Guo, Q. and Tang, L. (2015), An improved scatter search algorithm for the
  single machine total weighted tardiness scheduling problem with
  sequence-dependent setup times.
\newblock \emph{Applied Soft Computing}, v. , \penalty0 n. 0, p. \penalty0 --.
\newblock Forthcoming.

\bibitem[Gupta and Smith(2006)]{Gupta2006}
Gupta, S.~R. and Smith, J.~S. (2006), Algorithms for single machine total
  tardiness scheduling with sequence dependent setups.
\newblock \emph{European Journal of Operational Research}, v. 175, \penalty0 n.
  2, p. \penalty0 722--739.

\bibitem[Kirlik and O{\u g}uz(2012)]{Kirlik2012}
Kirlik, G. and O{\u g}uz, C. (2012), A variable neighborhood search for
  minimizing total weighted tardiness with sequence dependent setup times on a
  single machine.
\newblock \emph{Computers \& Operations Research}, v. 39, \penalty0 n. 7, p.
  \penalty0 1506--1520.

\bibitem[Lawler(1977)]{Lawler1977}
Lawler, E.~L.
\newblock A "pseudopolynomial" algorithm for sequencing jobs to minimize total
  tardiness.
\newblock P.L.~Hammer, E.L.~Johnson, B.~K. and Nemhauser, G. (Eds.),
  \emph{Studies in Integer Programming}, volume~1 of \emph{Annals of Discrete
  Mathematics}, p. 331--342. Elsevier, 1977.

\bibitem[Lee \emph{et~al}.(1997)Lee, Bhaskaran, and Pinedo]{Lee1997}
Lee, Y.~H., Bhaskaran, K. and Pinedo, M. (1997), A heuristic to minimize the
  total weighted tardiness with sequence-dependent setups.
\newblock \emph{IIE Transactions}, v. 29, \penalty0 n. 1, p. \penalty0 45--52.

\bibitem[Lenstra \emph{et~al}.(1977)Lenstra, Kan, and Brucker]{Lenstra1977}
Lenstra, J., Kan, A.~R. and Brucker, P.
\newblock Complexity of machine scheduling problems.
\newblock P.L.~Hammer, E.L.~Johnson, B.~K. and Nemhauser, G. (Eds.),
  \emph{Studies in Integer Programming}, volume~1 of \emph{Annals of Discrete
  Mathematics}, p. 343--362. Elsevier, 1977.

\bibitem[Liao and Cheng(2007)]{Liao2007}
Liao, C.-J. and Cheng, C.-C. (2007), A variable neighborhood search for
  minimizing single machine weighted earliness and tardiness with common due
  date.
\newblock \emph{Computers \& Industrial Engineering}, v. 52, \penalty0 n. 4, p.
  \penalty0 404--413.

\bibitem[Liao and Juan(2007)]{LiaoJuan2007}
Liao, C.-J. and Juan, H.-C. (2007), An ant colony optimization for
  single-machine tardiness scheduling with sequence-dependent setups.
\newblock \emph{Computers \& Operations Research}, v. 34, \penalty0 n. 7, p.
  \penalty0 1899--1909.

\bibitem[Liao \emph{et~al}.(2012)Liao, Tsou, and Huang]{Liaoetal2012}
Liao, C.-J., Tsou, H.-H. and Huang, K.-L. (2012), Neighborhood search
  procedures for single machine tardiness scheduling with sequence-dependent
  setups.
\newblock \emph{Theoretical Computer Science}, v. 434, p. \penalty0 45--52.

\bibitem[Lin and Ying(2007)]{Lin2007}
Lin, S.-W. and Ying, K.-C. (2007), Solving single-machine total weighted
  tardiness problems with sequence-dependent setup times by meta-heuristics.
\newblock \emph{The International Journal of Advanced Manufacturing
  Technology}, v. 34, \penalty0 n. 11-12, p. \penalty0 1183--1190.

\bibitem[Mandahawi \emph{et~al}.(2011)Mandahawi, Al-Shihabi, and
  Altarazi]{Mandahawi2011}
Mandahawi, N., Al-Shihabi, S. and Altarazi, S. (2011), A max-min ant system to
  minimize total tardiness on a single machine with sequence dependet setup
  times implementing a limited budget local search.
\newblock \emph{International Journal of Research \& Reviews in Applied
  Sciences}, v. 6.

\bibitem[Martin \emph{et~al}.(1991)Martin, Otto, and Felten]{Martin1991}
Martin, O., Otto, S.~W. and Felten, E.~W. (1991), Large-step markov chains for
  the traveling salesman problem.
\newblock \emph{Complex Systems}, v. 5, p. \penalty0 299--326.

\bibitem[Martinelli \emph{et~al}.(2013)Martinelli, Poggi, and
  Subramanian]{Martinellietal2013}
Martinelli, R., Poggi, M. and Subramanian, A. (2013), Improved bounds for large
  scale capacitated arc routing problem.
\newblock \emph{Computers \& Operations Research}, v. 40, \penalty0 n. 8, p.
  \penalty0 2145--2160.

\bibitem[Mladenovi{\'c} and Hansen(1997)]{MladenovicHansen1997}
Mladenovi{\'c}, N. and Hansen, P. (1997), Variable neighborhood search.
\newblock \emph{Computers \& Operations Research}, v. 24, \penalty0 n. 11, p.
  \penalty0 1097--1100.

\bibitem[Nawaz \emph{et~al}.(1983)Nawaz, Jr, and Ham]{Nawaz1983}
Nawaz, M., Jr, E. E.~E. and Ham, I. (1983), A heuristic algorithm for the
  m-machine, n-job flow-shop sequencing problem.
\newblock \emph{Omega}, v. 11, \penalty0 n. 1, p. \penalty0 91--95.

\bibitem[Penna \emph{et~al}.(2013)Penna, Subramanian, and Ochi]{Pennaetal2013}
Penna, P., Subramanian, A. and Ochi, L. (2013), An iterated local search
  heuristic for the heterogeneous fleet vehicle routing problem.
\newblock \emph{Journal of Heuristics}, v. 19, \penalty0 n. 2, p. \penalty0
  201--232.

\bibitem[Raman \emph{et~al}.(1989)Raman, Rachamadugu, and Talbot]{Raman1989}
Raman, N., Rachamadugu, R.~V. and Talbot, F. (1989), Real-time scheduling of an
  automated manufacturing center.
\newblock \emph{European Journal of Operational Research}, v. 40, \penalty0 n.
  2, p. \penalty0 222--242.

\bibitem[Rubin and Ragatz(1995)]{Rubin1995}
Rubin, P.~A. and Ragatz, G.~L. (1995), Scheduling in a sequence dependent setup
  environment with genetic search.
\newblock \emph{Computers \& Operations Research}, v. 22, \penalty0 n. 1, p.
  \penalty0 85--99.
\newblock Genetic Algorithms.

\bibitem[Silva \emph{et~al}.(2012)Silva, Subramanian, Vidal, and
  Ochi]{Silvaetal2012}
Silva, M.~M., Subramanian, A., Vidal, T. and Ochi, L.~S. (2012), A simple and
  effective metaheuristic for the minimum latency problem.
\newblock \emph{European Journal of Operational Research}, v. 221, \penalty0 n.
  3, p. \penalty0 513--520.

\bibitem[Subramanian(2012)]{Subramanian2012b}
Subramanian, A.
\newblock \emph{{Heuristic, Exact and Hybrid Approaches for Vehicle Routing
  Problems}}.
\newblock PhD thesis, Universidade Federal Fluminense, Niter\'{o}i, Brazil,
  2012.

\bibitem[Subramanian and Battarra(2013)]{SubramanianBattarra2013}
Subramanian, A. and Battarra, M. (2013), An iterated local search algorithm for
  the travelling salesman problem with pickups and deliveries.
\newblock \emph{Journal of the Operational Research Society}, v. 64, \penalty0
  n. 3, p. \penalty0 402--409.

\bibitem[Subramanian \emph{et~al}.(2010)Subramanian, Drummond, Bentes, Ochi,
  and Farias]{Subramanian2010}
Subramanian, A., Drummond, L., Bentes, C., Ochi, L. and Farias, R. (2010), A
  parallel heuristic for the vehicle routing problem with simultaneous pickup
  and delivery.
\newblock \emph{Computers \& Operations Research}, v. 37, \penalty0 n. 11, p.
  \penalty0 1899--1911.

\bibitem[Subramanian \emph{et~al}.(2014)Subramanian, Battarra, and
  Potts]{Subramanian2014}
Subramanian, A., Battarra, M. and Potts, C.~N. (2014), An iterated local search
  heuristic for the single machine total weighted tardiness scheduling problem
  with sequence-dependent setup times.
\newblock \emph{International Journal of Production Research}, v. 52, \penalty0
  n. 9, p. \penalty0 2729--2742.

\bibitem[Tanaka and Araki(2013)]{Tanaka2013}
Tanaka, S. and Araki, M. (2013), An exact algorithm for the single-machine
  total weighted tardiness problem with sequence-dependent setup times.
\newblock \emph{Computers \& Operations Research}, v. 40, \penalty0 n. 1, p.
  \penalty0 344--352.

\bibitem[Tasgetiren \emph{et~al}.(2009)Tasgetiren, Pan, and
  Liang]{Tasgetiren2009}
Tasgetiren, M.~F., Pan, Q.-K. and Liang, Y.-C. (2009), A discrete differential
  evolution algorithm for the single machine total weighted tardiness problem
  with sequence dependent setup times.
\newblock \emph{Computers \& Operations Research}, v. 36, \penalty0 n. 6, p.
  \penalty0 1900--1915.

\bibitem[Valente and Alves(2008)]{Valente2008}
Valente, J.~M. and Alves, R.~A. (2008), Beam search algorithms for the single
  machine total weighted tardiness scheduling problem with sequence-dependent
  setups.
\newblock \emph{Computers \& Operations Research}, v. 35, \penalty0 n. 7, p.
  \penalty0 2388--2405.

\bibitem[Vidal \emph{et~al}.(2015)Vidal, Battarra, Subramanian, and Erdo{\v
  g}an]{Vidaletal2015}
Vidal, T., Battarra, M., Subramanian, A. and Erdo{\v g}an, G. (2015), Hybrid
  metaheuristics for the clustered vehicle routing problem.
\newblock \emph{Computers \& Operations Research.}, v. 58, p. \penalty0 87 --
  99.

\bibitem[Xu \emph{et~al}.(2013)Xu, L\"{u}, and Cheng]{Xu2013}
Xu, H., L\"{u}, Z. and Cheng, T. (2013), Iterated local search for
  single-machine scheduling with sequence-dependent setup times to minimize
  total weighted tardiness.
\newblock \emph{Journal of Scheduling}, v. 17, p. \penalty0 271--287.

\bibitem[Xu \emph{et~al}.(2014)Xu, L{\" u}, Yin, Shen, and Buscher]{Xu2014}
Xu, H., L{\" u}, Z., Yin, A., Shen, L. and Buscher, U. (2014), A study of
  hybrid evolutionary algorithms for single machine scheduling problem with
  sequence-dependent setup times.
\newblock \emph{Computers \& Operations Research}, v. 50, p. \penalty0 47 --
  60.

\bibitem[Ying \emph{et~al}.(2009)Ying, Lin, and Huang]{Ying2009}
Ying, K.-C., Lin, S.-W. and Huang, C.-Y. (2009), Sequencing single-machine
  tardiness problems with sequence dependent setup times using an iterated
  greedy heuristic.
\newblock \emph{Expert Systems with Applications}, v. 36, \penalty0 n. 3, Part
  2, p. \penalty0 7087--709.

\end{thebibliography}










\end{document}